\documentclass{article}
\pdfoutput=1

\PassOptionsToPackage{numbers}{natbib}
\usepackage{titletoc}
\usepackage{hyperref}
\usepackage[page,header]{appendix}
\usepackage[final]{nips_2017}
\usepackage[utf8]{inputenc} 
\usepackage[T1]{fontenc}    
\usepackage{url}            
\usepackage{booktabs}       
\usepackage{amsfonts}       
\usepackage{nicefrac}       
\usepackage{microtype}      
\usepackage{xcolor}
\usepackage{hyperref}
\usepackage{graphicx}
\usepackage{algorithmicx}
\usepackage{algorithm}
\usepackage{algpseudocode}
\usepackage{mathrsfs}
\usepackage{amssymb}
\usepackage{graphicx,wrapfig,lipsum}
\usepackage{natbib}
\usepackage{caption}
\usepackage{float}
\usepackage{subcaption}
\usepackage{enumitem}
\usepackage{siunitx}
\usepackage{multirow}
\usepackage{tikz}
\usepackage{footnote}
\usetikzlibrary{arrows,shapes,snakes,automata,backgrounds,petri,positioning,calc}
\usepackage{lipsum}
\usepackage[multiple]{footmisc}
\usepackage{placeins}
\renewcommand{\S}{\mathcal{S}}

\newcommand{\A}{\mathcal{A}}

\newcommand{\R}{\mathbb{R}}
\newcommand{\E}{\mathbb{E}}

\newif\ifshort
\shortfalse

\ifshort
    \usepackage{titlesec}
    \titlespacing\section{0pt}{0pt plus 4pt minus 2pt}{0pt plus 2pt minus 2pt}
    \titlespacing\subsection{0pt}{0pt plus 4pt minus 2pt}{0pt plus 2pt minus 2pt}
\else
    
\fi

\title{Learning Dexterous In-Hand Manipulation}

\author{\\
\textbf{OpenAI}\thanks{Please use the following bibtex for citation: \url{https://openai.com/bibtex/openai2018learning.bib}},  Marcin~Andrychowicz, Bowen~Baker, Maciek~Chociej,\\
Rafał~Józefowicz, Bob~McGrew, Jakub~Pachocki, Arthur~Petron,\\
Matthias~Plappert, Glenn~Powell, Alex~Ray, Jonas~Schneider, Szymon~Sidor, \\
Josh~Tobin, Peter~Welinder, Lilian~Weng, Wojciech~Zaremba
}

\begin{document}
\maketitle

\begin{figure}[h!]
\centering
\includegraphics[width=1.0\textwidth]{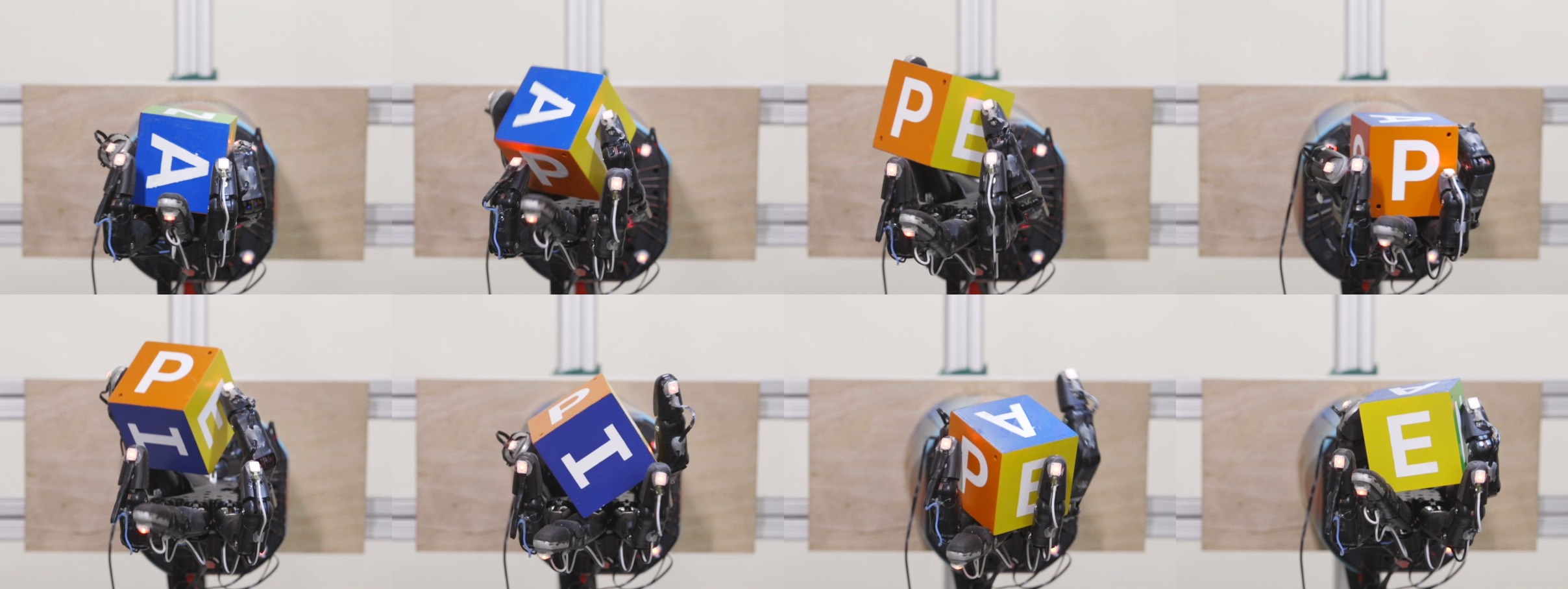} \\ \vspace{0.3cm}
\includegraphics[height=1cm]{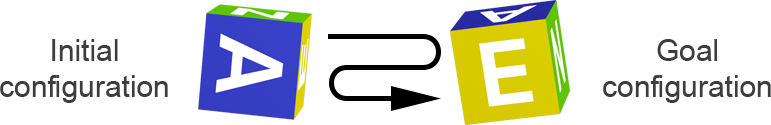} \\ \vspace{0.1cm}
\caption{A five-fingered humanoid hand trained with reinforcement learning manipulating a block from an initial configuration to a goal configuration using vision for sensing.}
\end{figure}

\begin{abstract}
We use reinforcement learning (RL) to learn dexterous in-hand manipulation
policies which can perform vision-based object reorientation on a physical
Shadow Dexterous Hand.
The training is performed in a simulated environment
in which we randomize many of the physical properties of the system like
friction coefficients and an object's appearance.
Our policies transfer to the physical robot
despite being trained entirely in simulation.
Our method does not rely on any human demonstrations, but many behaviors found in human manipulation emerge naturally, including finger gaiting, multi-finger coordination, and
the controlled use of gravity.
Our results were obtained using the same distributed RL system
that was used to train OpenAI~Five~\citep{five}.
We also include a video of our results: \url{https://youtu.be/jwSbzNHGflM}.
\end{abstract}

\startcontents[mainsections]

\section{Introduction}

\begin{figure}\centering
    \includegraphics[width=\textwidth]{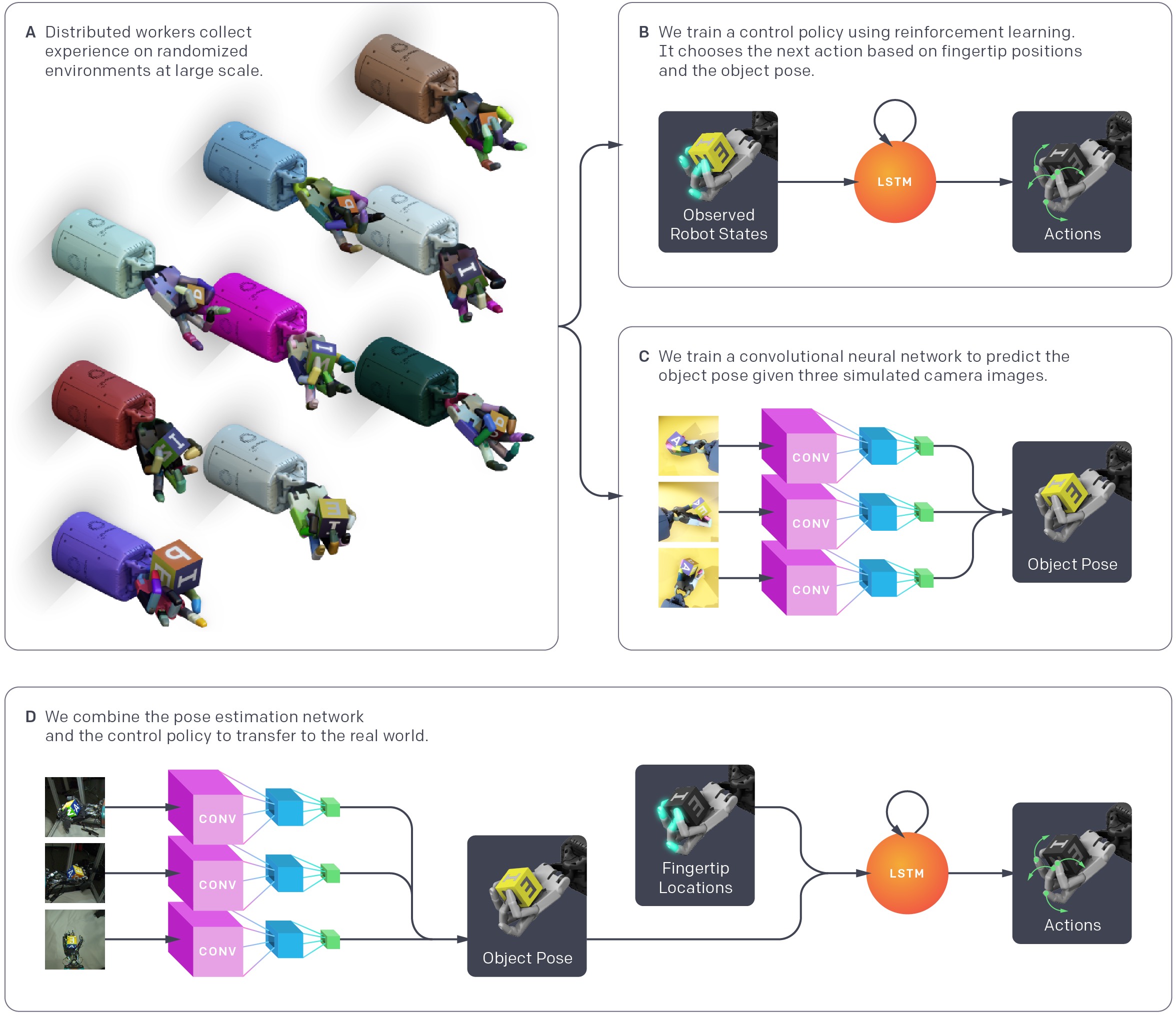}
    \caption{
        System Overview. (a) We use a large distribution of simulations with randomized parameters and appearances to collect data for both the control policy and vision-based pose estimator. (b) The control policy receives observed robot states and rewards from the distributed simulations and learns to map observations to actions using a recurrent neural network and reinforcement learning. (c) The vision based pose estimator renders scenes collected from the distributed simulations and learns to predict the pose of the object from images using a convolutional neural network (CNN), trained separately from the control policy. (d) To transfer to the real world, we predict the object pose from 3 real camera feeds with the CNN, measure the robot fingertip locations using a 3D motion capture system, and give both of these to the control policy to produce an action for the robot.
    }
    \label{fig:overview}
\end{figure}

While dexterous manipulation of objects is a fundamental everyday task for humans,
it is still challenging for autonomous robots.
Modern-day robots are typically designed for specific tasks in constrained settings and are largely unable to utilize complex end-effectors.
In contrast, people are able to perform a wide range of dexterous manipulation tasks in a diverse set of environments, making the human hand a grounded source of inspiration for research into robotic manipulation.

The Shadow Dexterous Hand~\citep{shadow-robot} is an example of a robotic hand designed for human-level dexterity; it has five fingers with a total of \num{24} degrees of freedom.
The hand has been commercially available since 2005; however it still has not seen widespread adoption, which can be attributed to the daunting difficulty of controlling systems of such complexity.
The state-of-the-art in controlling five-fingered hands is severely limited.
Some prior methods have shown promising in-hand manipulation results in simulation but do not attempt to transfer to a real world robot \citep{DBLP:conf/icra/BaiL14, DBLP:conf/sca/MordatchPT12}.
Conversely, due to the difficulty in modeling such complex systems, there has also been work in approaches that only train on a physical robot \citep{falco2018policy, DBLP:conf/humanoids/HoofHN015, DBLP:journals/corr/KumarGTL16, DBLP:conf/icra/KumarTL16}.
However, because physical trials are so slow and costly to run, the learned behaviors are very limited.

In this work, we demonstrate methods to train control policies that perform in-hand manipulation 
and deploy them on a physical robot.
The resulting policy exhibits unprecedented levels of dexterity and naturally discovers grasp types found in humans, such as the tripod, prismatic, and tip pinch grasps, 
and displays contact-rich, dynamic behaviours such as finger gaiting, multi-finger coordination, the controlled use of gravity, and coordinated application of translational and torsional forces to the object.
Our policy can also use vision to sense an object's pose --- an important aspect for robots that should ultimately work outside of a controlled lab setting.

Despite training entirely in a simulator which substantially differs from the real world,
we obtain control policies which perform well on the physical robot.
We attribute our transfer results to (1) extensive randomizations and added effects in the simulated environment alongside calibration, (2) memory augmented control polices which admit the possibility to learn adaptive behaviour and implicit system identification on the fly, and (3) training at large scale with distributed reinforcement learning.
An overview of our approach is depicted in \autoref{fig:overview}.

The paper is structured as follows.
\autoref{sec:setup} gives a system overview, describes the proposed task in more detail, and shows the hardware setup. \autoref{sec:randomizations} describes observations for the control policy, environment randomizations, and additional effects added to the simulator that make transfer possible.
\autoref{sec:train-policy} outlines the control policy training procedure and the distributed RL system.
\autoref{sec:train-vision} describes the vision model architecture and training procedure.
Finally, \autoref{sec:results} describes both qualitative and quantitative results from deploying the control policy and vision model on a physical robot. 

\section{Task and System Overview}
\label{sec:setup}

\begin{figure}
    \centering
    \begin{subfigure}[b]{\textwidth}
        \centering
        \includegraphics[width=\textwidth]{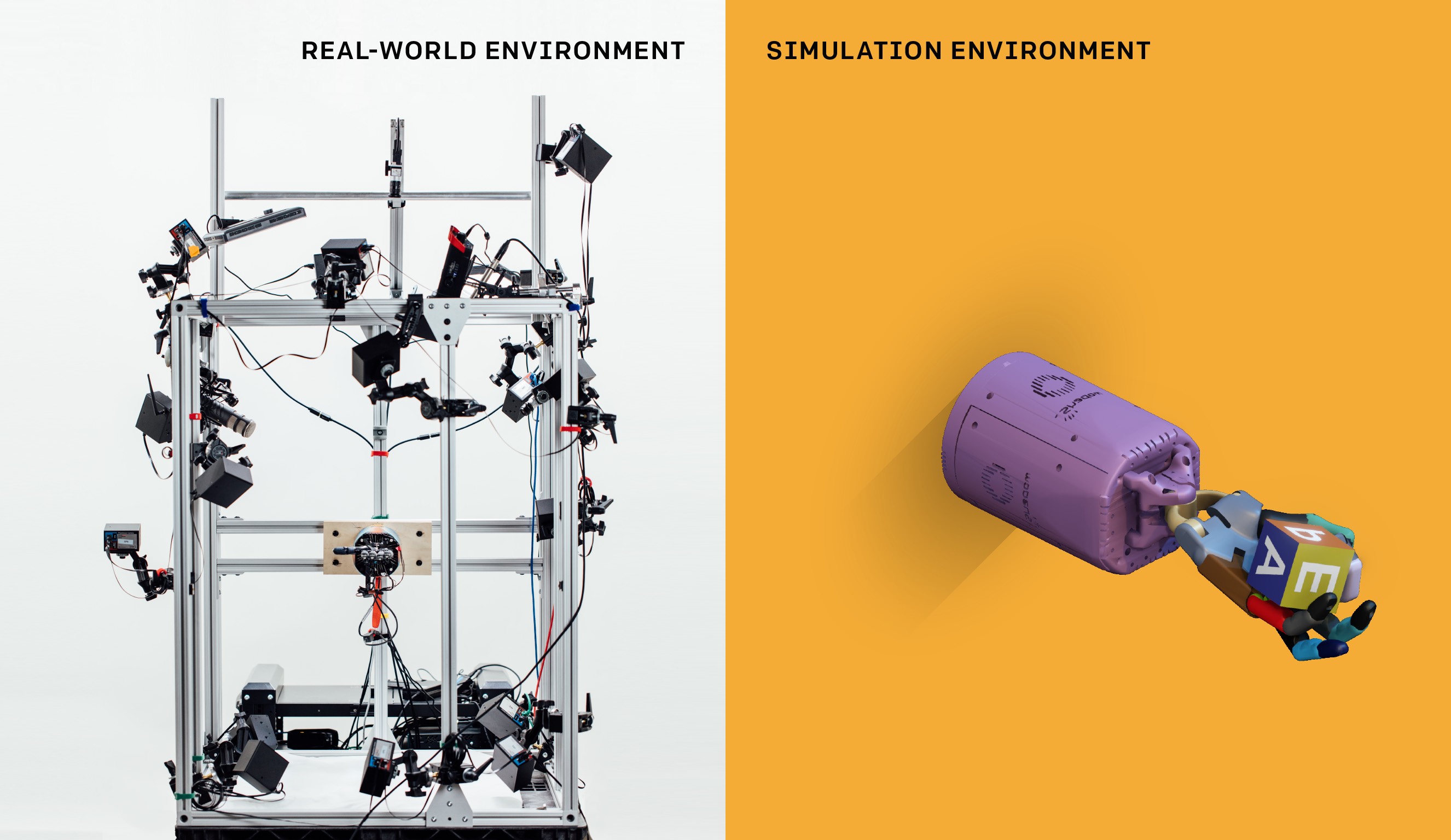}
    \end{subfigure}%
    \caption{(left) The "cage" which houses the robot hand, 16 PhaseSpace tracking cameras, and 3 Basler RGB cameras. (right) A rendering of the simulated environment.
    }
    \label{fig:setup}
\end{figure}

In this work we consider the problem of in-hand object reorientation.
We place the object under consideration onto the palm of a humanoid robot hand.
The goal is to reorient the object to a desired target configuration in-hand.
As soon as the current goal is (approximately) achieved, a new goal is provided until the object is eventually dropped.
We use two different objects, a block and an octagonal prism.
\autoref{fig:setup} depicts our physical system as well as our simulated environment.

\subsection{Hardware}

We use the Shadow Dexterous Hand, which is a humanoid robotic hand with $24$ degrees of freedom (DoF) actuated by $20$ pairs of agonist--antagonist tendons.
We use a PhaseSpace motion capture system to track the Cartesian position of all five finger tips.
For the object pose, we have two setups: One that uses PhaseSpace markers to track the object and one that uses three Basler RGB cameras for vision-based pose estimation.
This is because our goal is to eventually have a system that works outside of a lab environment, and vision-based systems are better equipped to handle the real world.
We do not use the touch sensors embedded in the hand and only use joint sensing for implementing low-level relative position control. We update the targets of the low level controller, which runs at roughly \SI{1}{kHz}, with relative positions given by the control policy at roughly \SI{12}{Hz}.

More details on our hardware setup are available in \autoref{app:hardware}.

\subsection{Simulation}
We simulate the physical system with the MuJoCo physics engine~\citep{MuJoCo}, and we use Unity\footnote{Unity game engine website: \url{https://unity3d.com/}} to render the images
for training the vision based pose estimator. Our model of the Shadow Dexterous Hand is based on the one used in the OpenAI~Gym robotics environments~\citep{plappert2018multi} but has been improved to match the physical system more closely through calibration (see Appendix~\ref{app:model-calibration} for details).

Despite our calibration efforts, the simulation is still a rough approximation of the physical setup.
For example, our model directly applies torque to joints instead of tendon-based actuation and uses rigid body contact models instead of deformable body contact models.
Modeling these and other effects seen in the real world is difficult or impossible in a rigid body simulator. 
These differences cause a "reality gap" and make it unlikely for a policy trained in a simulation with these inaccuracies to transfer well.

We describe additional details of our simulation in Appendix~\ref{app:sim}.

\section{Transferable Simulations}
\label{sec:randomizations}

As described in the previous section, our simulation is a coarse approximation of the real world.
We therefore face a dilemma: we cannot train on the physical robot because deep reinforcement learning algorithms require millions of samples;
conversely, training only in simulation results in policies that do no transfer well due to the gap between the simulated and real environments.
To overcome the reality gap, we modify the basic version of our simulation to a \emph{distribution over many simulations} that foster transfer~\citep{DBLP:conf/rss/SadeghiL17, tobin2017domain, peng2017sim}.
By carefully selecting the sensing modalities and by randomizing most aspects of our simulated environment we are able to train policies that are less likely to overfit to a specific simulated environment and more likely to transfer successfully to the physical robot.

\subsection{Observations}

We give the control policy observations of the fingertips using PhaseSpace markers and the object pose either from PhaseSpace markers or the vision based pose estimator.
Although the Shadow Dexterous Hand contains a broad array of built-in sensors, we specifically avoided providing these as observations to the policy because they are subject to state-dependent noise that would have been difficult to model in the simulator.
For example, the fingertip tactile sensor measures the pressure of a fluid stored in a balloon inside the fingertip, which correlates with the force applied to the fingertip but also with a number of confounding variables, including atmospheric pressure, temperature, and the shape of the contact and intersection geometry.
Although it is straightforward to determine the existence of contacts in the simulator, it would be difficult to model the distribution of sensor values.
Similar considerations apply to the joint angles measured by Hall effect sensors, which are used by the low-level controllers but not provided to the policy due to their tendency to be noisy and hard to calibrate.

\subsection{Randomizations}
\label{subsection:randomizations}

Following previous work on \emph{domain randomization}~\citep{DBLP:conf/rss/SadeghiL17, tobin2017domain, peng2017sim},
we randomize most of the aspects of the simulated environment in order to learn both a policy and a vision model that generalizes to reality. We briefly detail the types of randomizations below, and Appendix~\ref{app:randomizations} contains a more detailed discussion of the more involved randomizations and provides hyperparameters. 

\paragraph{Observation noise.}
To better mimic the kind of noise we expect to experience in reality, we add Gaussian noise to policy observations.
In particular, we apply a correlated noise which is sampled once per episode
as well as an uncorrelated noise sampled at every timestep.

\paragraph{Physics randomizations.} Physical parameters like friction are randomized at the beginning of every episode and held fixed. Many parameters are centered on values found during model calibration in an effort to make the simulation distribution match reality more closely. \autoref{table:rand-physics} lists all physics parameters that are randomized.

\begin{table}
    \footnotesize
    \centering
    \caption{Ranges of physics parameter randomizations.}
    \renewcommand{\arraystretch}{1.3}
    \begin{tabular}{@{}lll@{}}
        \toprule
        \textbf{Parameter} & \textbf{Scaling factor range} & \textbf{Additive term range} \\ \midrule
        object dimensions & $\mbox{uniform}([0.95,1.05])$ & \\
        object and robot link masses & $\mbox{uniform}([0.5,1.5])$ & \\
        surface friction coefficients & $\mbox{uniform}([0.7,1.3])$ & \\
        robot joint damping coefficients & $\mbox{loguniform}([0.3,3.0])$ & \\
        actuator force gains (P term) & $\mbox{loguniform}([0.75,1.5])$ & \\ \hline
        joint limits & & $\mathcal{N}(0,0.15)~\si{\radian}$  \\
        gravity vector (each coordinate) && $\mathcal{N}(0,0.4)~\si{\m\per\s^2}$ \\ 
        \bottomrule
    \end{tabular}
\label{table:rand-physics}
\end{table}

\paragraph{Unmodeled effects.}
The physical robot experiences many effects that are not modeled by our simulation.
To account for imperfect actuation, we use a simple model of motor backlash and introduce action delays and action noise before applying them in simulation.
Our motion capture setup sometimes loses track of a marker temporarily, which we model by freezing the position of a simulated marker with low probability for a short period of time in simulation.
We also simulate marker occlusion by freezing its simulated position whenever
it is close to another marker or the object.
To handle additional unmodeled dynamics, we apply small random forces to the object.
Details on the concrete implementation are available in Appendix~\ref{app:randomizations}.

\begin{figure}[h]
    \begin{center}
    \includegraphics[width=\textwidth]{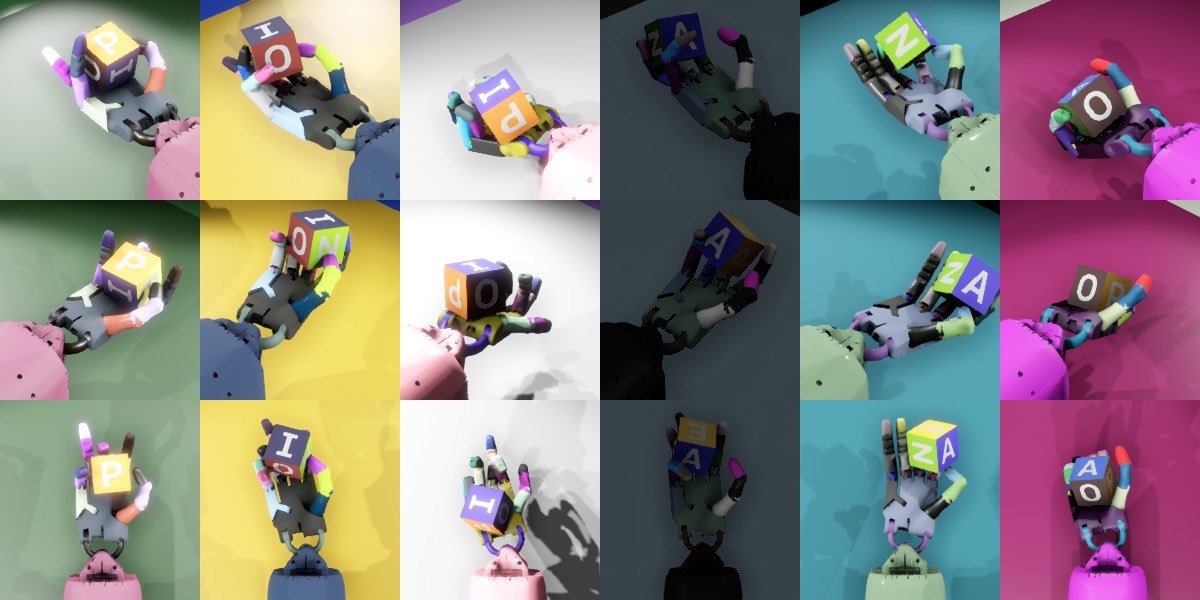}
    \caption{Simulations with different randomized visual appearances. Rows correspond to the renderings from the same camera, and columns correspond to renderings from 3 separate cameras which are simultaneously fed into the neural network.}
    \label{fig:random_img}    
    \end{center}
\end{figure}

\paragraph{Visual appearance randomizations.}
 We randomize the following aspects of the rendered scene:
camera positions and intrinsics, lighting conditions, the pose of the hand and object, and the materials and textures for all objects in the scene.
\autoref{fig:random_img} depicts some examples of these randomized environments.
Details on the randomized properties and their ranges are available in Appendix~\ref{app:randomizations}.

\section{Learning Control Policies From State}
\label{sec:train-policy}

\subsection{Policy Architecture}

Many of the randomizations we employ persist across an episode, and thus it should be possible for a memory augmented policy to identify properties of the current environment and adapt its own behavior accordingly.
For instance, initial steps of interaction with the environment can reveal the weight of the object or how fast the index finger can move.
We therefore represent the policy as a recurrent neural network with memory, namely an LSTM~\citep{lstm}
with an additional hidden layer with ReLU~\citep{relu} activations inserted
between inputs and the LSTM.

The policy is trained with Proximal Policy Optimiztion (PPO)~\citep{ppo}.
We provide background on reinforcement learning and PPO in greater detail in~\autoref{sec:rl}.
PPO requires the training of two networks --- a policy network, which maps observations to actions, and a value network, which predicts the discounted sum of future rewards starting from a given state.
Both networks have the same architecture but have independent parameters.
Since the value network is only used during training, we use an Asymmetric Actor-Critic~\citep{pinto2017asymmetric} approach.
Asymmetric Actor-Critic exploits the fact that the value network can have access to information that is not available on the real robot system.\footnote{This includes noiseless observation and additional observations like joint angles and angular velocities, which we cannot sense reliably but which are readily available in simulation during training.}
This potentially simplifies the problem of learning good value estimates since less information needs to be inferred.
The list of inputs fed to both networks can be found in \autoref{table:policy-inputs}.

\makesavenoteenv{table}
\makesavenoteenv{tabular}
\begin{table}[h!]
    \footnotesize
    \centering
    \caption{Observations of the policy and value networks, respectively.}
    \renewcommand{\arraystretch}{1.3}
    \begin{tabular}{@{}llcc@{}}
        \toprule
        \textbf{Input} & \textbf{Dimensionality} & \textbf{Policy network} & \textbf{Value network} \\
        \midrule
        fingertip positions & 15D & \checkmark & \checkmark \\
        object position & 3D & \checkmark & \checkmark \\
        object orientation & 4D (quaternion) & $\times$\footnote{We accidentally did not include the current object orientation in the policy observations but found that it makes little difference since this information is indirectly available through the relative target orientation.} & \checkmark \\
        target orientation & 4D (quaternion) & $\times$ & \checkmark \\
        relative target orientation & 4D (quaternion) & \checkmark & \checkmark \\
        hand joints angles & 24D & $\times$ & \checkmark \\
        hand joints velocities & 24D & $\times$ & \checkmark \\
        object velocity & 3D & $\times$ & \checkmark \\
        object angular velocity & 4D (quaternion) & $\times$ & \checkmark \\
        \bottomrule
    \end{tabular}
\label{table:policy-inputs}
\end{table}

\newcommand{\link}[5]{
    \draw[->,flow,#3] ([xshift=-5pt]#1.south) -- node[left]{#4} ([xshift=-5pt]#2.north);
    \draw[->,flow,#3] ([xshift=5pt]#2.north) -- node[right]{#5} ([xshift=5pt]#1.south);
}

\newcommand{\optthread}[2]{
    \begin{scope}
      \node [thread,#2] (train#1) {optimizer};
      \node [physical,below=0.5cm of train#1] (gpu#1) {GPU};
      \node [thread,below=0.5cm of gpu#1] (stager#1) {stager};
      \node [physical,below=0.5cm of stager#1] (ram#1) {RAM};
      \node [thread,below=0.5cm of ram#1] (puller#1) {puller};
      \link{train#1}{gpu#1}{}{}{}
      \link{gpu#1}{stager#1}{}{}{}
      \link{stager#1}{ram#1}{}{}{}
      \link{ram#1}{puller#1}{}{}{}
    \end{scope}
}

\newcommand{\redis}[1]{
    \node[db,below=1.5cm of puller#1] (redis#1) {redis};
    \link{puller#1}{redis#1}{dashed}{parameters}{experience}
      
    \node[thread,below=1.5cm of redis#1] (w1) {worker};
    \node[thread,below right=0.1cm and 0.1cm of w1.north west] (w2) {worker};
    \node[thread,below right=0.1cm and 0.1cm of w2.north west] (w3) {worker};
    \node[thread,below right=0.1cm and 0.1cm of w3.north west] (w4) {worker};
    \node[thread,below right=0.1cm and 0.1cm of w4.north west] (w5) {worker};
    \link{redis#1}{w1}{dashed}{parameters}{experience}
      
}

\subsection{Actions and Rewards}

Policy actions correspond to desired joints angles relative to the current ones\footnote{The reason
for using \emph{relative} targets is that it is hard to precisely measure absolute joints angles
on the physical robot. See Appendix~\ref{app:hardware} for more details.}
(e.g. rotate this joint by $10$ degrees).
While PPO can handle both continuous and discrete action spaces, we noticed that
discrete action spaces work much better. This may be because a discrete probability distribution is more expressive than a multivariate Gaussian or because discretization of actions makes learning a good advantage function potentially simpler.
We discretize each action coordinate into $11$ bins.

The reward given at timestep $t$ is $r_t=d_t-d_{t+1}$, where
$d_t$ and $d_{t+1}$ are the rotation angles between the desired 
and current object orientations before and after the transition, respectively.
We give an additional reward of $5$ whenever a goal is achieved and a reward of $-20$ (a penalty) whenever the object is dropped.
More information about the simulation environment
can be found in Appendix~\ref{app:sim}.

\subsection{Distributed Training with Rapid}
\begin{figure}[h]
    \centering
    \includegraphics[width=0.7\textwidth]{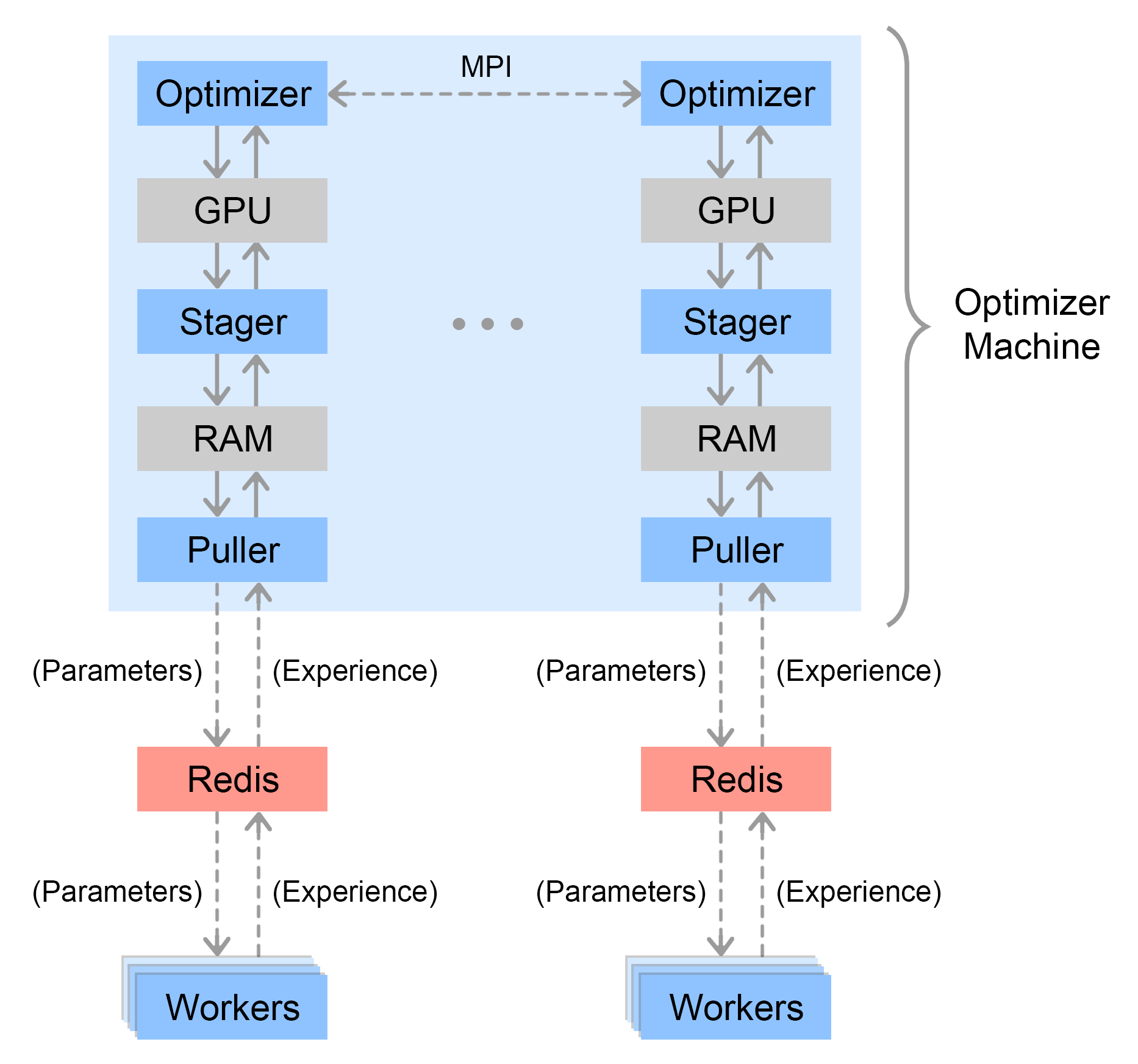}
    \caption{Our distributed training infrastructure in Rapid. Individual threads are depicted as blue squares. Worker machines randomly connect to a Redis server from which they pull new policy parameters and to which they send new experience. The optimizer machine has one MPI process for each GPU, each of which gets a dedicated Redis server. Each process has a \emph{Puller} thread which pulls down new experience from Redis into a buffer. Each process also has a \emph{Stager} thread which samples minibatches from the buffer and stages them on the GPU. Finally, each \emph{Optimizer} thread uses a GPU to optimize over a minibatch after which gradients are accumulated across threads and new parameters are sent to the Redis servers.}
    \label{fig:rapid}
\end{figure}

We use the same distributed implementation of PPO that was used to train OpenAI Five~\citep{five} without any modifications.
Overall, we found that PPO scales up easily and requires little hyperparameter tuning. 
The architecture of our distributed training system is depicted in~\autoref{fig:rapid}.

In our implementation, a pool of $384$ worker machines, each with $16$ CPU cores, generate experience by rolling out the current version of the policy in a sample from distribution of randomized simulations.
Workers download the newest policy parameters from the optimizer at the beginning of every epoch,
generate training episodes, and send the generated episodes back to the optimizer.
The communication between the optimizer and workers is implemented using the
Redis in-memory data store.
We use multiple Redis instances for load-balancing, and workers are assigned to an instance randomly.
This setup allows us to generate about $2$ years of simulated experience per hour.

The optimization is performed on a single machine with $8$ GPUs.
The optimizer threads pull down generated experience from Redis
and then stage it to their respective GPU's memory for processing.
After computing gradients locally, they are averaged across all threads using MPI, which we then use to update
the network parameters.

The hyperparameters that we used can be found in Appendix~\ref{app:hyper-ppo}.

\section{State Estimation from Vision}
\label{sec:train-vision}
The policy that we describe in the previous section
takes the object's position as input and requires a motion capture system for tracking the object
on the physical robot.
This is undesirable because tracking objects with such a system is only feasible in a lab setting where markers can be placed on each object.
Since our ultimate goal is to build robots for the real world that can interact with arbitrary objects, sensing them using vision is an important building block.
In this work, we therefore wish to infer the object's pose from vision alone.
Similar to the policy, we train this estimator only on synthetic data coming from the simulator. 

\subsection{Model Architecture}
\label{sec:vision_model_arch}

\begin{figure}[h]
    \begin{minipage}[c]{0.45\textwidth}
        \includegraphics[width=0.8\textwidth]{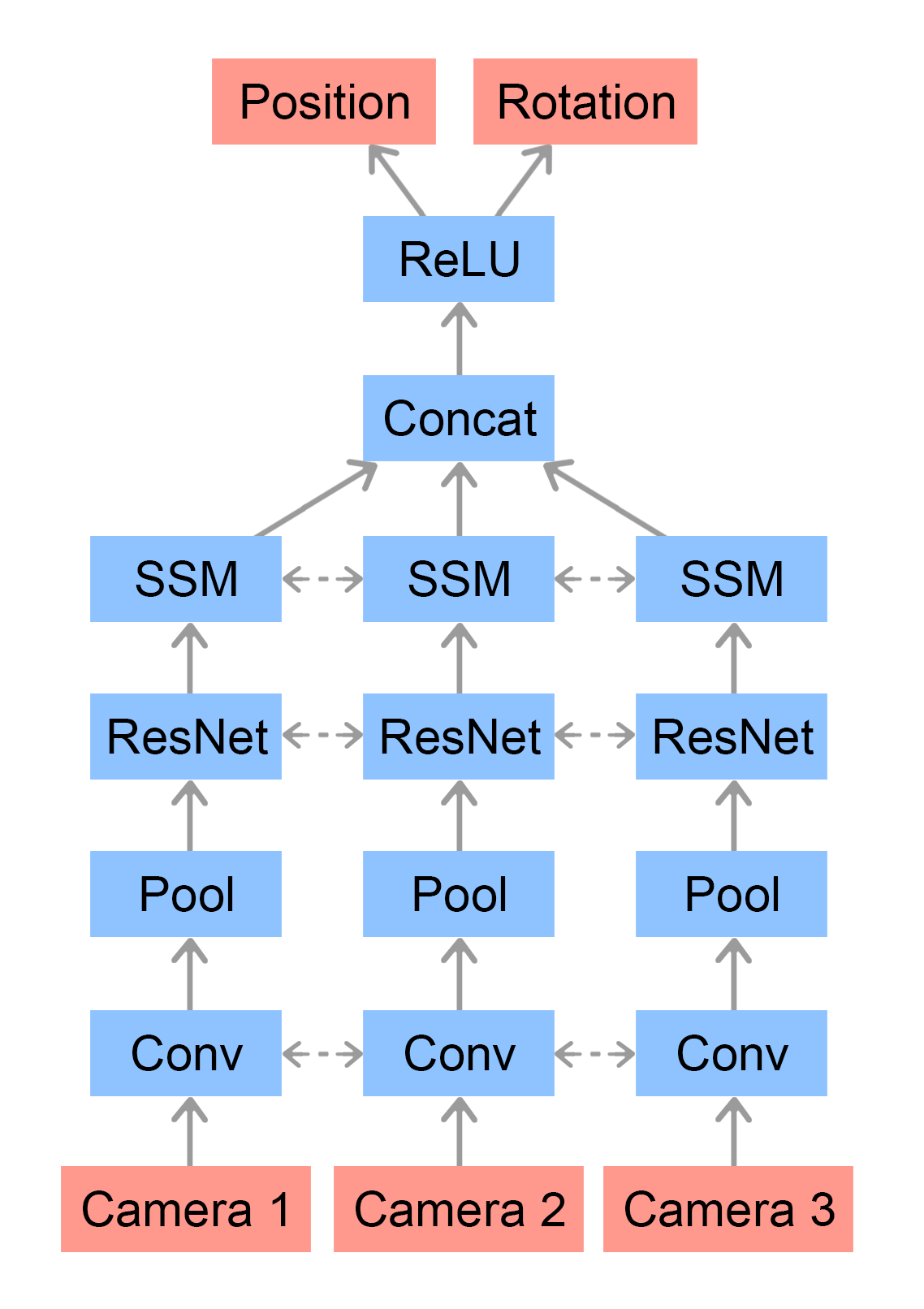}
    \end{minipage}\hfill
    \begin{minipage}[c]{0.55\textwidth}
        \caption{Vision network architecture. Camera images are passed through a convolutional feature stack that consists of two convolutional layers, max-pooling, 4 ResNet blocks~\cite{He2016DeepRL}, and spatial softmax (SSM)~\cite{finn2015deep} layers with shared weights between the feature stacks for each camera.
        The resulting representations are flattened, concatenated, and fed to a fully connected network. 
        All layers use ReLU \cite{relu} activation function.
        Linear outputs from the last layer form the estimates of the position and orientation of the object.
        }
        \label{fig:vision-architecture}
    \end{minipage}\hfill
\end{figure}

To resolve ambiguities and to increase robustness, we use three RGB cameras mounted with differing viewpoints of the scene.
The recorded images are passed through a convolutional neural network, which is depicted in \autoref{fig:vision-architecture}.
The network predicts both the position and the orientation of the object.
During execution of the control policy on the physical robot, we feed the pose estimator's prediction into the policy,
which in turn produces the next action.

\subsection{Training}

We run the trained policy in the simulator until we gather one million states.
We then train the vision network by minimizing the mean squared error between the normalized prediction and the ground-truth
with minibatch gradient descent.
For each minibatch, 
we render the images with randomized appearance before feeding them to the network.
Moreover, we augment the data by modifying the object pose.
We use 2 GPUs for rendering and 1 GPU for running the network and training.

Additional training details are available in Appendix~\ref{app:vision_training} and randomization details are in Appendix~\ref{app:randomizations}.

\section{Results}
\label{sec:results}
In this section, we evaluate the proposed system.
We start by deploying the system on the physical robot,
and evaluating its performance on in-hand manipulation of a block and an octagonal prism.
We then focus on individual aspects of our system:
We conduct an ablation study of the importance of randomizations
and policies with memory capabilities in order to successfully transfer.
Next, we consider the sample complexity of our proposed method. 
Finally, we investigate the performance of the proposed vision pose estimator and show that using only synthetic images is sufficient to achieve good performance.

\subsection{Qualitative Results}

\begin{figure}[h]
    \centering
    \includegraphics[width=0.32\textwidth]{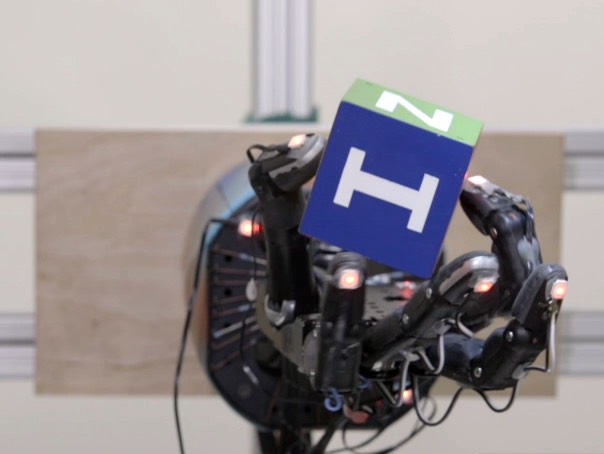}
    \includegraphics[width=0.32\textwidth]{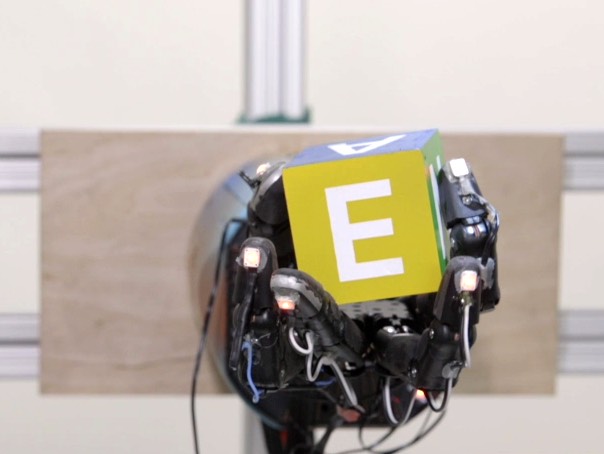}
    \includegraphics[width=0.32\textwidth]{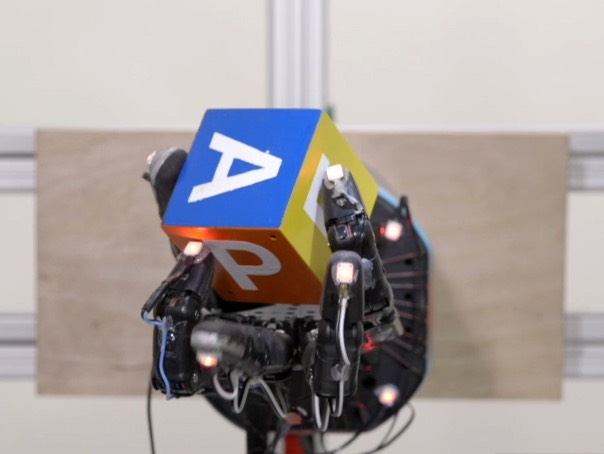}\\ \vspace{0.05cm}
    \includegraphics[width=0.32\textwidth]{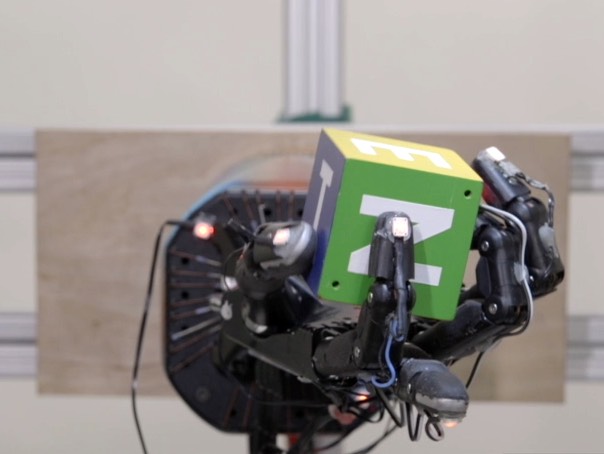}
    \includegraphics[width=0.32\textwidth]{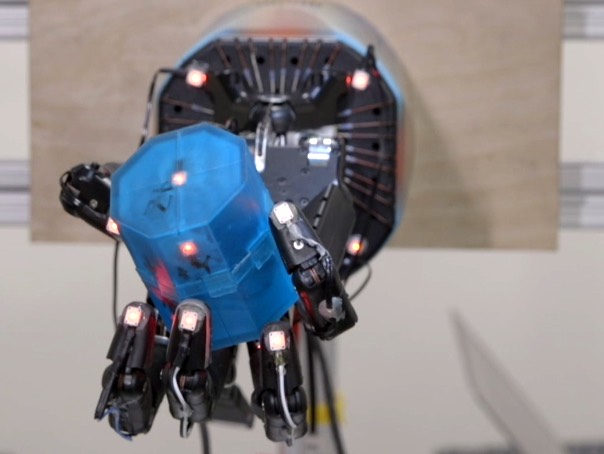}
    \includegraphics[width=0.32\textwidth]{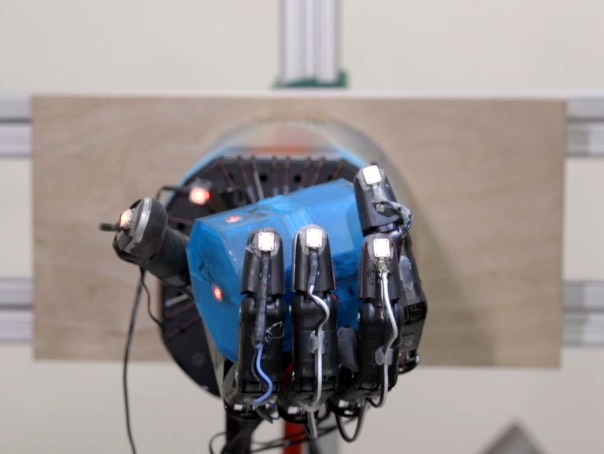}
    \caption{Different grasp types learned by our policy. From top left to bottom right: Tip Pinch grasp, Palmar Pinch grasp, Tripod grasp, Quadpod grasp, 5-Finger Precision grasp, and a Power grasp. Classified according to~\citep{grasp}.}
    \label{fig:grasps}
\end{figure}

During deployment on the robot as well as in simulation, we notice that our policies naturally exhibit many of the grasps found in humans (see \autoref{fig:grasps}).
Furthermore, the policy also naturally discovers many strategies for dexterous in-hand manipulation described by the robotics community~\citep{DBLP:conf/icar/MaD11} such as finger pivoting, finger gaiting,
multi-finger coordination, the controlled use of gravity, and coordinated application of translational and torsional forces to the object.
It is important to note that we did not incentivize this directly: we do not use any human demonstrations and do not encode any prior into the reward function.

For precision grasps, our policy tends to use the little finger instead of the index or middle finger.
This may be because the little finger of the Shadow Dexterous Hand has an extra degree of freedom compared to the index, middle and ring fingers, making it more dexterous.
In humans the index and middle finger are typically more dexterous.
This means that our system can rediscover grasps found in humans, but adapt them to better fit the limitations and abilities of its own body.

We observe another interesting parallel between humans and our policy in finger pivoting, which is a strategy in which an object is held between two fingers and rotate around this axis.
It was found that young children have not yet fully developed their motor skills and therefore tend to rotate objects using the proximal or middle phalanges of a finger~\citep{pehoski1997hand}.
Only later in their lives do they switch to primarily using the distal phalanx, which is the dominant strategy found in adults.
It is interesting that our policy also typically relies on the distal phalanx for finger pivoting.

During experiments on the physical robot we noticed that the most common failure mode was dropping the object
while rotating the wrist pitch joint down.
Moreover, the vertical joint was the most common source of robot breakages, probably because
it handles the biggest load.
Given these difficulties,
we also trained a policy with the wrist pitch joint locked.\footnote{We had trouble training
in this environment from scratch, so we fine-tuned
a policy trained in the original environment instead.}
We noticed that not only does this policy transfer better to the physical robot but
it also seems to handle the object much more deliberately with many of the above grasps emerging frequently in this setting.
Other failure modes that we observed were dropping the object shortly after the start of a trial (which may be explained by incorrectly identifying some aspect of the environment) and getting stuck because the edge of an object got caught in a screw hole (which we do not model).

We encourage the reader to watch the accompanying video to get a better sense of the learned behaviors.\footnote{Real-time video of $50$ successful consecutive rotations: \url{https://youtu.be/DKe8FumoD4E}}

\subsection{Quantitative Results}
\label{sec:quant-results}
In this section we evaluate our results quantitatively.
To do so, we measure the number of \emph{consecutive} successful rotations until the object is either dropped, a goal has not been achieved within 80 seconds, or until $50$ rotations are achieved.
All results are available in~\autoref{table:perf}.

\begin{table}[h]
    \centering
    \renewcommand{\arraystretch}{1.1}
    \caption{The number of successful consecutive rotations in simulation and on the physical robot. All policies were trained on environments with all randomizations enabled. We performed 100 trials in simulation and 10 trails per policy on the physical robot. Each trial terminates when the object is dropped, 50 rotations are achieved or a timeout is reached. For physical trials, results were taken at different times on the physical robot.}
    \begin{tabular}{@{}llll@{}}
        \toprule
        \textbf{Simulated task} & \textbf{Mean} & \textbf{Median} & \textbf{Individual trials (sorted)} \\ 
        \midrule
        Block (state) & $43.4 \pm 13.8$ & $50$ & - \\
        Block (state, locked wrist) & $44.2 \pm 13.4$ & $50$ & - \\
        Block (vision) & $30.0 \pm 10.3$ & $33$ & - \\
        Octagonal prism (state) & $29.0 \pm 19.7$ & $30$ & - \\
        \midrule
        \textbf{Physical task} \\
        \midrule

        Block (state) & $18.8 \pm 17.1$  & $13$ & $50$, $41$, $29$, $27$, $14$, $12$, $6$, $4$, $4$, $1$ \\
        
        Block (state, locked wrist) & $26.4 \pm 13.4$ & $28.5$ & $50$, $43$, $32$, $29$, $29$, $28$, $19$, $13$, $12$, $9$ \\
        
        Block (vision) & $15.2 \pm 14.3$ & $11.5$ & $46$, $28$, $26$, $15$, $13$, $10$, $8$, $3$, $2$, $1$ \\
        Octagonal prism (state) & $7.8 \pm 7.8$ & $5$ & $27$, $15$, $8$, $8$, $5$, $5$, $4$, $3$, $2$, $1$ \\
        \bottomrule
    \end{tabular}
    \label{table:perf}
\end{table}

Our results allow us to directly compare the performance of each task in simulation and on the real robot.
For instance, manipulating a block in simulation achieves a median of $50$ successes while the median on the physical setup is $13$.
This is the overall trend that we observe: Even though randomizations and calibration narrow the reality gap, it still exists and performance on the real system is worse than in simulation. We discuss the importance of individual randomizations in greater detail in \autoref{sec:ablation-rand}.

When using vision for pose estimation, we achieve slightly worse results both in simulation and on the real robot. This is because even in simulation, our model has to perform transfer because it was only trained on images rendered with Unity but we use MuJoCo rendering for evaluation in simulation (thus making this a sim-to-sim transfer problem).
On the real robot, our vision model does slightly worse compared to pose estimation with PhaseSpace. However, the difference is surprisingly small, suggesting that training the vision model only in simulation is enough to achieve good performance on the real robot.
For vision pose estimation we found that it helps to use a white background and to wipe the object with a tack cloth between trials to remove detritus from the robot hand.

We also evaluate the performance on a second type of object, an octagonal prism.
To do so, we finetuned a trained block rotation control policy to the same randomized distribution of environments but with the octagonal prism as the target object instead of the block.
Even though our randomizations were all originally designed for the block, we were able to learn successful policies that transfer.
Compared to the block, however, there is still a performance gap both in simulation and on the real robot.
This suggests that further tuning is necessary and that the introduction of additional randomization could improve transfer to the physical system.

We also briefly experimented with a sphere but failed to achieve more than a few rotations in a row, perhaps
because we did not randomize any MuJoCo parameters related to rolling behavior or because rolling objects are much more sensitive to unmodeled imperfections in the hand such as screw holes.
It would also be interesting to train a unified policy that can handle multiple objects, but we leave this for future work.

Obtaining the results in \autoref{table:perf} proved to be challenging due to robot breakages during experiments.
Repairing the robot takes time and often changes some aspects of the system, which is why the results were obtained at different times.
In general, we found that problems with hardware breakage were one of the key challenges we had to overcome in this work.

\subsection{Ablation of Randomizations}
\label{sec:ablation-rand}

\begin{figure}[h]
    \begin{minipage}[c]{0.55\textwidth}
        \includegraphics[width=0.95\textwidth]{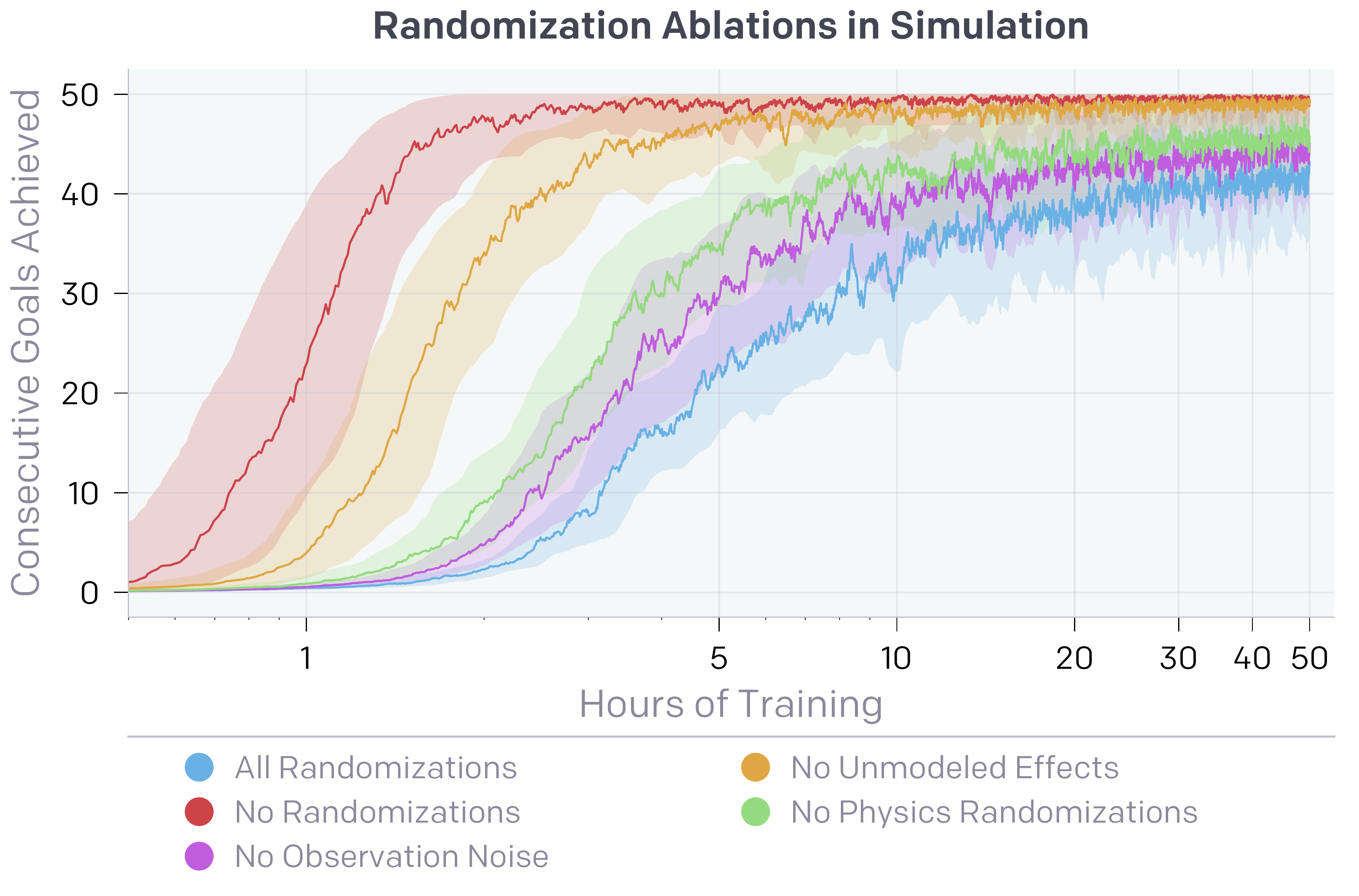}
    \end{minipage}\hfill
    \begin{minipage}[c]{0.45\textwidth}
        \caption{Performance when training in environments with groups of randomizations held out. All runs show exponential moving averaged performance and 90\% confidence interval over a moving window of the RL agent in the environment it was trained. We see that training is faster in environments that are easier, e.g. \textit{no randomizations} and \textit{no unmodeled effects}. We only show one seed per experiment; however, in general we have noticed almost no instability in training.
        }
        \label{fig:rand-abl}
    \end{minipage}\hfill
\end{figure}

In \autoref{subsection:randomizations} we detail a list of parameters we randomize and effects we add that are not already modeled in the simulator. In this section we show that these additions to the simulator are vital for transfer. We train 5 separate RL policies in environments with various randomizations held out: \textit{all randomizations} (baseline), \textit{no observation noise}, \textit{no unmodeled effects}, \textit{no physics randomizations}, and \textit{no randomizations} (basic simulator, i.e. no domain randomization).

Adding randomizations or effects to the simulation does not come without cost; in \autoref{fig:rand-abl} we show the training performance in simulation for each environment plotted over wall-clock time. Policies trained in environments with a more difficult set of randomizations, e.g. \textit{all randomizations} and \textit{no observation noise}, converge much slower and therefore require more compute and simulated experience to train in. However, when deploying these policies on the real robot we find that training with randomizations is critical for transfer. \autoref{table:rand-abl} summarizes our results.
Specifically, we find that training with all randomizations leads to a median of $13$ consecutive goals achieved, while policies trained with \textit{no randomizations}, \textit{no physics randomizations}, and \textit{no unmodeled effects} achieve only median of $0$, $2$, and $2$ consecutive goals, respectively.

\begin{table}[h]
    \centering
    \caption{
    The number of successful consecutive rotations on the physical robot of 5 policies trained separately in environments with different randomizations held out.
    The first 5 rows use PhaseSpace for object pose estimation and were run on the same robot at the same time. Trials for each row were interleaved in case the state of the robot changed during the trials. The last two rows were measured at a different time from the first 5 and used the vision model to estimate the object pose.
    }
    \renewcommand{\arraystretch}{1.3}
    \begin{tabular}{@{}llll@{}}
        \toprule
        \textbf{Training environment} & \textbf{Mean} & \textbf{Median} & \textbf{Individual trials (sorted)} \\ 
        \midrule
        All randomizations (state) & $18.8 \pm 17.1$  & $13$ & $50$, $41$, $29$, $27$, $14$, $12$, $6$, $4$, $4$, $1$ \\
        No randomizations (state) & $1.1 \pm 1.9$ & $0$ & $6$, $2$, $2$, $1$, $0$, $0$, $0$, $0$, $0$, $0$ \\
        No observation noise (state) & $15.1 \pm 14.5$ & $8.5$ & $45$, $35$, $23$, $11$, $9$, $8$, $7$, $6$, $6$, $1$ \\
        No physics randomizations (state) & $3.5 \pm 2.5$ & $2$ & $7$, $7$, $7$, $3$, $2$, $2$, $2$, $2$, $2$, $1$ \\
        No unmodeled effects (state) & $3.5 \pm 4.8$ & $2$ & $16$, $7$, $3$, $3$, $2$, $2$, $1$, $1$, $0$, $0$ \\
        \midrule
        All randomizations (vision) & $15.2 \pm 14.3$ & $11.5$ & $46$, $28$, $26$, $15$, $13$, $10$, $8$, $3$, $2$, $1$ \\
        No observation noise (vision) & $5.9 \pm 6.6$ & $3.5$ & $20$, $12$, $11$, $6$, $5$, $2$, $2$, $1$, $0$, $0$ \\
    \bottomrule
    \end{tabular}
    \label{table:rand-abl}
\end{table}

When holding out \emph{observation noise} randomizations, the performance gap is less clear than for the other randomization groups.
We believe that is because our motion capture system has very little noise.
However, we still include this randomization because it is important when the vision and control policies are composed.
In this case, the pose estimate of the object is much more noisy, and, therefore, training with observation noise should be more important.
The results in \autoref{table:rand-abl} suggest that this is indeed the case, with a drop from median performance of $11.5$ to $3.5$ if the observation noise randomizations are withheld.

The vast majority of training time is spent making the policy robust to different physical dynamics. Learning to rotate an object in simulation without randomizations requires about 3 years of simulated experience, while achieving the same performance in a fully randomized simulation requires about 100 years of experience.
This corresponds to a wall-clock time of around 1.5 hours and 50 hours in our simulation setup, respectively.

\subsection{Effect of Memory in Policies}

\begin{figure}[h]
    \begin{minipage}[c]{0.55\textwidth}
        \includegraphics[width=0.95\textwidth]{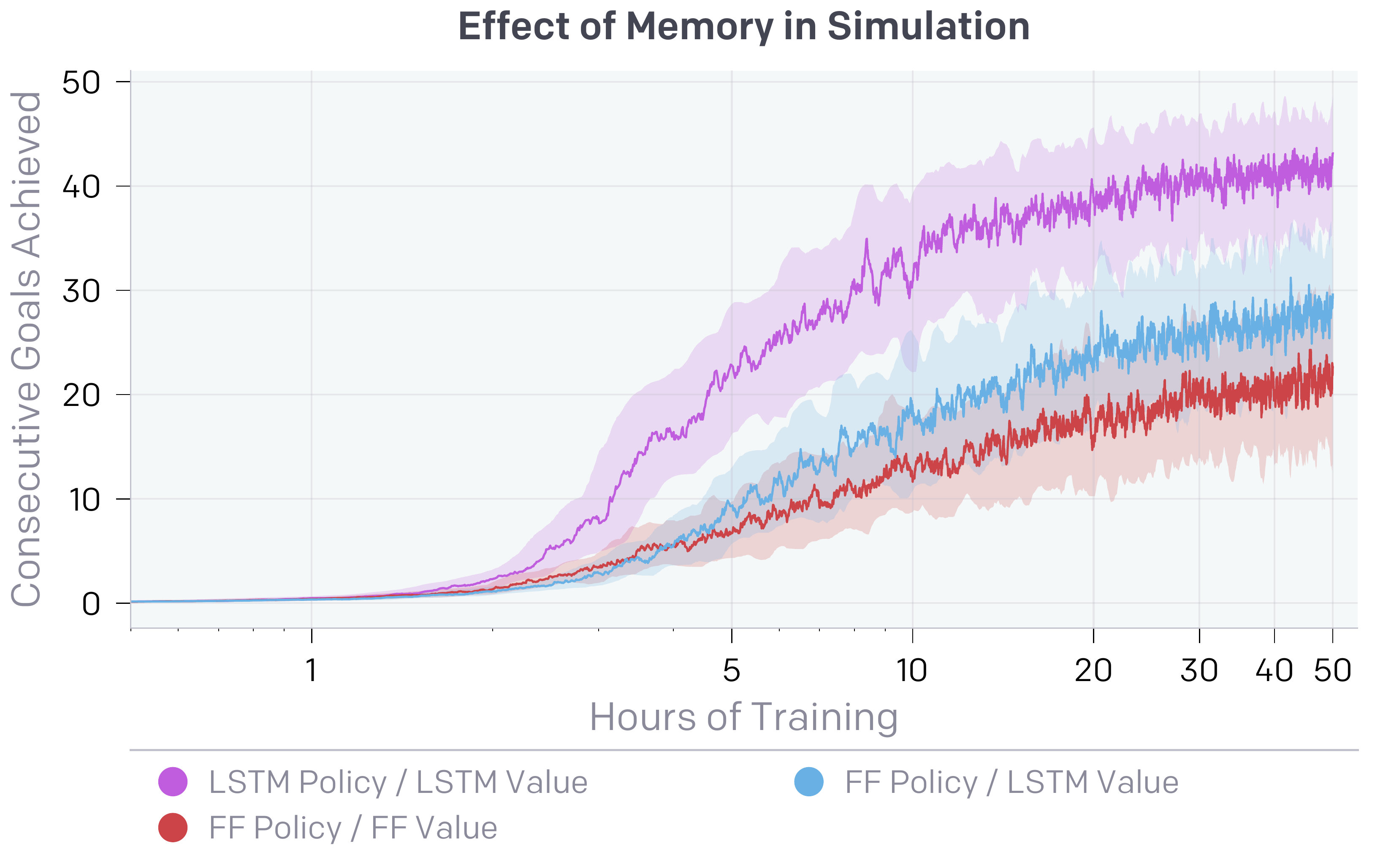}
    \end{minipage}\hfill
    \begin{minipage}[c]{0.45\textwidth}
        \caption{Performance when comparing LSTM and feed forward (FF) policy and value function networks. We train on an environment with all randomizations enabled. All runs show exponential moving averaged performance and 90\% confidence interval over a moving window for a single seed. We find that using recurrence in both the policy and value function helps to achieve good performance in simulation.
        }
        \label{fig:mem-abl}
    \end{minipage}\hfill
\end{figure}

We find that using memory is helpful to achieve good performance in the randomized simulation. In \autoref{fig:mem-abl} we show the simulation performance of three different RL architectures: the baseline which has a LSTM policy and value function, a feed forward (FF) policy and a LSTM value function, and both a  FF policy and FF value function. We include results for a FF policy with LSTM value function because it was plausible that having a more expressive value function would accelerate training, allowing the policy to act robustly without memory once it converged. However, we see that the baseline outperforms both variants, indicating that it is beneficial to have some amount of memory in the actual policy.

Moreover, we found out that LSTM state is predictive of the environment randomizations.
In particular, we discovered that the LSTM hidden state after 5 seconds of simulated interaction
with the block allows to predict whether the block is bigger or smaller than
average in $80\%$ of cases.

To investigate the importance of memory-augmented policies for transfer, we evaluate the same three network architectures as described above on the physical robot. \autoref{table:memory-abl} summarizes the results.
Our results show that having a policy with access to memory yields a higher median of successful rotations, suggesting that the policy may use memory to adapt to the current environment.\footnote{When training in an environment with no randomizations, the FF and LSTM policy converge to the same performance in the same amount of time. This shows that a FF policy has the capacity and observations to solve the non-randomized task but cannot solve it reliably with all randomizations, plausibly because it cannot adapt to the environment.}
Qualitatively we also find that FF policies often get stuck and then run out of time.

\begin{table}[h]
    \centering
    \caption{
    The number of successful consecutive rotations on the physical robot of 3 policies with different network architectures trained on an environment with all randomizations.
    Results for each row were collected at different times on the physical robot.
    }
    \renewcommand{\arraystretch}{1.3}
    \begin{tabular}{@{}llll@{}}
        \toprule
        \textbf{Network architecture} & \textbf{Mean} & \textbf{Median} & \textbf{Individual trials (sorted)} \\ 
        \midrule
        LSTM policy / LSTM value (state) & $18.8 \pm 17.1$  & $13$ & $50$, $41$, $29$, $27$, $14$, $12$, $6$, $4$, $4$, $1$ \\
        FF policy / LSTM value (state) & $4.7 \pm 4.1$ & $3.5$ & $15$, $7$, $6$, $5$, $4$, $3$, $3$, $2$, $2$, $0$ \\
        FF policy / FF value (state) & $4.6 \pm 4.3$ & $3$ & $15$, $8$, $6$, $5$, $3$, $3$, $2$, $2$, $2$, $0$ \\
    \bottomrule
    \end{tabular}
    \label{table:memory-abl}
\end{table}

\subsection{Sample Complexity \& Scale}

\begin{figure}[h!]
    \centering
    \begin{subfigure}[t]{0.45\textwidth}
        \centering
        \includegraphics[width=\textwidth]{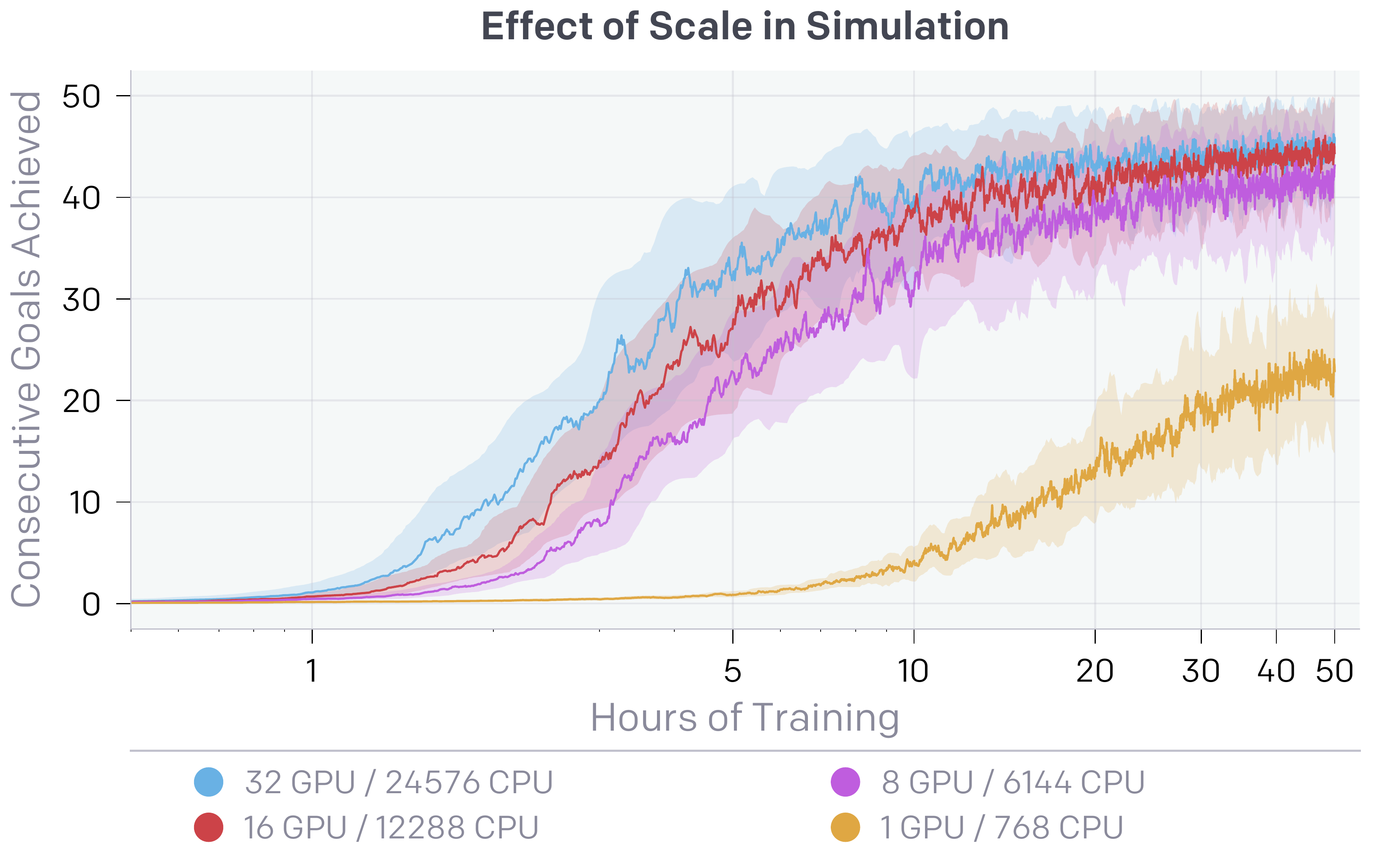}
    \end{subfigure}
    \begin{subfigure}[t]{0.45\textwidth}
        \centering
        \includegraphics[width=\textwidth]{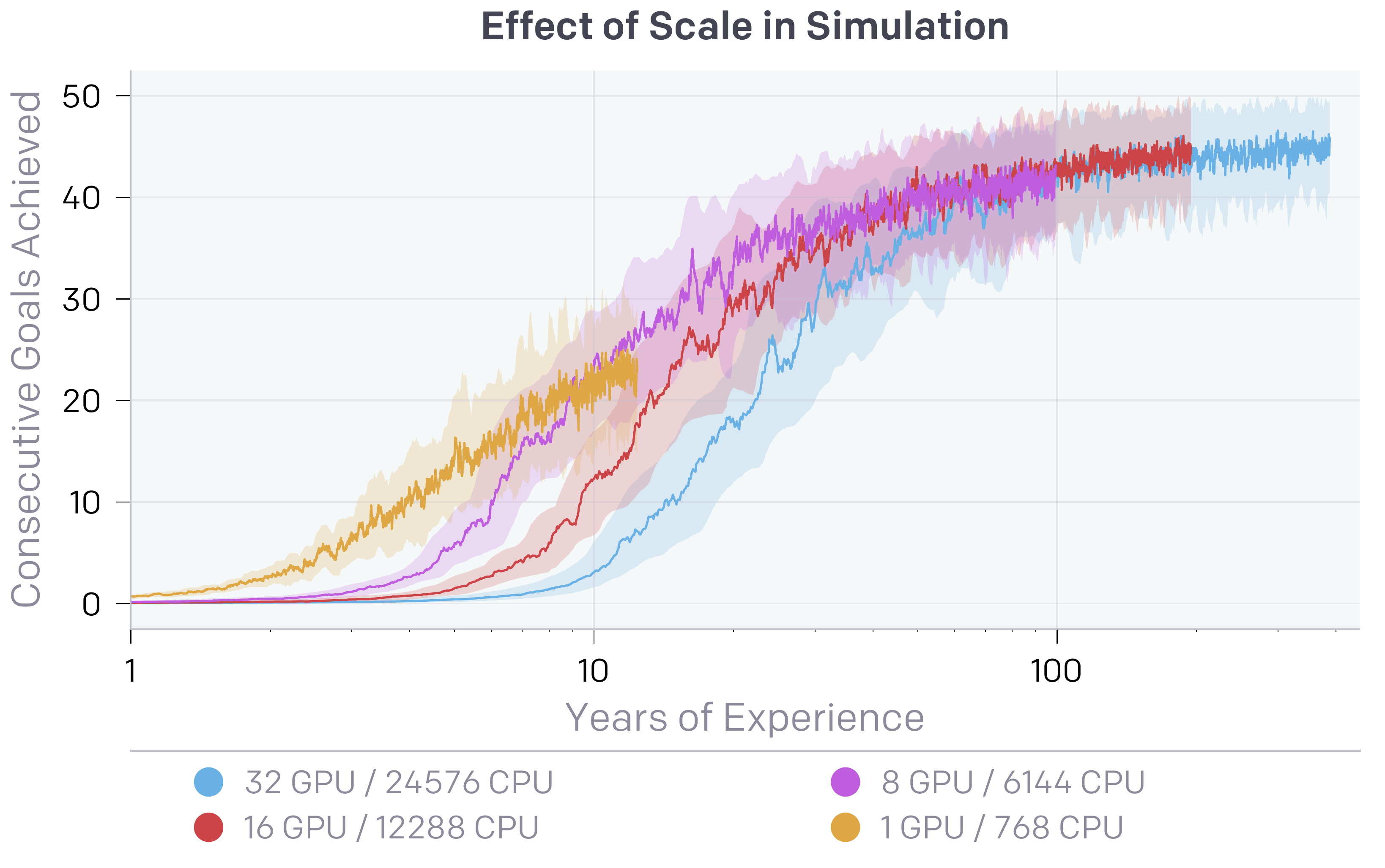}
    \end{subfigure}
    \caption{We show performance in simulation when varying the amount of compute used during training versus wall clock training time (left) and years of experience consumed (right). Batch size used is proportional to the number of GPUs used, such that time per optimization step should remain constant apart from slow downs due to gradient syncing across optimizer machines.}
    \label{fig:scale}
\end{figure}

In \autoref{fig:scale} we show results when varying the number of CPU cores and GPUs used in training, where we keep the batch size per GPU fixed such that overall batch size is directly proportional to number of GPUs.
Because we could linearly slow down training by simply using less CPU machines and having the GPUs wait longer for data, it is more informative to vary the batch size.
We see that our default setup with an 8 GPU optimizer and 6144 rollout CPU cores
reaches 20 consecutive achieved goals approximately 5.5 times faster than a setup with a 1 GPU optimizer and 768 rollout cores. Furthermore, when using 16 GPUs we reach 40 consecutive achieved goals roughly 1.8 times faster than when using the default 8 GPU setup.
Scaling up further results in diminishing returns, but it seems that scaling up to 16 GPUs and 12288 CPU cores gives close to linear speedup.

\subsection{Vision Performance}
\label{sec:result-vision}
In \autoref{table:perf} we show that we can combine a vision-based pose estimator and the control policy to successfully transfer to the real robot without embedding sensors in the target object.
To better understand why this is possible, we evaluate the precision of the pose estimator on both synthetic and real data.
Evaluating the system in simulation is easy because we can generate the necessary data and have access to the precise object's pose to compare against.
In contrast, real images had to be collected by running a state-based policy on our robot platform.
We use PhaseSpace to estimate the object's pose, which is therefore subject to errors.
The resulting collected test set consists of $992$ real samples.\footnote{A sample contains 3 images of the same scene. We removed a few samples that had no object in them after it being dropped.}
For simulation, we use test sets rendered using Unity and MuJoCo. The MuJoCo renderer was not used during training, thus the evaluation can be also considered as an instance of sim-to-sim transfer.
\autoref{table:vision} summarizes our results.

\begin{table}[h]
    \caption{Performance of a vision based pose estimator on synthetic and real data.}
    \centering
    \renewcommand{\arraystretch}{1.3}
    \begin{tabular}{@{}lll@{}}
        \toprule
        \textbf{Test set} & \textbf{Rotation error} & \textbf{Position error} \\ 
        \midrule
        Rendered images (Unity) & $2.71^{\circ} \pm 1.62$ & $3.12\text{mm} \pm 1.52$ \\
        Rendered images (MuJoCo) & $3.23^{\circ} \pm 2.91$ & $3.71\text{mm} \pm 4.07$ \\
        Real images & $5.01^{\circ} \pm 2.47$ & $9.27\text{mm} \pm 4.02$ \\
    \bottomrule\end{tabular}
    \label{table:vision}
\end{table}

Our results show that the model achieves low error for both rotation and position prediction when tested on synthetic data.\footnote{For comparison, PhaseSpace is rated for a position accuracy of around $20$ $\mu$m but requires markers and a complex setup.}
On the images rendered with MuJoCo, there is only a slight increase in error, suggesting successful sim-to-sim transfer.
The error further increases on the real data, which is due to the gap between simulation and reality but also because the ground truth is more challenging to obtain due to noise, occlusions, imperfect marker placement, and delayed sensor readings.
Despite that the prediction error is bigger than the observation noise used during policy training (\autoref{table:obs-noise}), the vision-based policy performs well
on the physical robot (\autoref{table:perf}).

\section{Related Work}
In order to make it easier to understand the state-of-the-art in dexterous in-hand manipulation
we gathered a representative set of videos from related work, and created a playlist\footnote{Related work playlist: \url{https://bit.ly/2uOK21Q}} out of them.

\subsection{Dexterous Manipulation}

Dexterous manipulation has been an active area of research for decades~\citep{DBLP:conf/icra/Fearing86, DBLP:journals/ijrr/Rus99,DBLP:journals/trob/Bicchi00, DBLP:conf/icra/OkamuraSC00, DBLP:conf/icar/MaD11}.
Many different approaches and strategies have been proposed over the years.
This includes rolling~\citep{DBLP:conf/icra/BicchiS95, DBLP:conf/icra/HanGLQT97, DBLP:conf/icra/HanT98, DBLP:journals/trob/CherifG99, DBLP:conf/icra/DoulgeriD13}, sliding~\citep{DBLP:journals/trob/CherifG99, DBLP:journals/trob/ShiWUL17}, finger gaiting~\citep{DBLP:conf/icra/HanT98}, finger tracking~\citep{DBLP:conf/icra/Rus92}, pushing~\citep{DBLP:journals/corr/DafleR17}, and re-grasping~\citep{DBLP:conf/icra/TournassoudLM87, DBLP:conf/icra/DafleRPTSEMLSF14}.
For some hand types, strategies like pivoting~\citep{DBLP:conf/iros/AiyamaII93}, tilting~\citep{DBLP:journals/trob/ErdmannM88}, tumbling~\citep{sawasaki1991tumbling}, tapping~\citep{DBLP:journals/ijrr/HuangM00}, two-point manipulation~\citep{DBLP:conf/iros/AbellE95}, and two-palm manipulation~\citep{DBLP:journals/ijrr/Erdmann98} are also options.
These approaches use planning and therefore require exact models of both the hand and object.
After computing a trajectory, the plan is typically executed open-loop, thus making these methods prone to failure if the model is not accurate.\footnote{Some methods use iterative re-planning to partially mitigate this issue.}

Other approaches take a closed-loop approach to dexterous manipulation and incorporate sensor feedback during execution, e.g. tactile sensing~\citep{DBLP:conf/icra/TaharaAY10, DBLP:conf/iros/LiBKB14, DBLP:conf/icra/LiYTB14, DBLP:journals/ijma/0001MHRB13}.
While those approaches allow to correct for mistakes at runtime, they still require reasonable models of the robot kinematics and dynamics, which can be challenging to obtain for under-actuated hands with many degrees of freedom.

Deep reinforcement learning has also been used successfully to learn complex manipulation skills on physical robots.
Guided policy search~\citep{DBLP:conf/icml/LevineK13, DBLP:conf/icra/LevineWA15} learns simple local policies directly on the robot and distills them into a global policy represented by a neural network.
An alternative is to use many physical robots simultaneously in order to be able to collect sufficient experience~\citep{DBLP:conf/icra/GuHLL17, DBLP:journals/ijrr/LevinePKIQ18, 2018arXiv180610293K}.

\subsection{Dexterous In-Hand Manipulation}
Since a very large body of past work on dexterous manipulation exists, we limit the more detailed discussion to setups that are most closely related to our work on dexterous in-hand manipulation.

Mordatch et al.~\citep{DBLP:conf/sca/MordatchPT12} and Bai et al.~\citep{DBLP:conf/icra/BaiL14} propose methods to generate trajectories for complex and dynamic in-hand manipulation, but results are limited to simulation. 
There has also been significant progress in learning complex in-hand dexterous manipulation~\citep{plappert2018multi, DBLP:journals/corr/abs-1804-08617} and even tool use~\citep{DBLP:journals/corr/abs-1709-10087} using deep reinforcement learning but those approaches were only evaluated in simulation as well.

In contrast, multiple authors learn policies for dexterous in-hand manipulation directly on the robot.
Hoof et al.~\citep{DBLP:conf/humanoids/HoofHN015} learn in-hand manipulation for a simple 3-fingered gripper whereas Kumar et al.~\citep{DBLP:conf/icra/KumarTL16, DBLP:journals/corr/KumarGTL16} and Falco et al.~\citep{falco2018policy} learn such policies for more complex humanoid hands.
While learning directly on the robot means that modeling the system is not an issue, it also means that learning has to be performed with only a handful of trials.
This is only possible when learning very simple (e.g. linear or local) policies that, in turn, do not exhibit sophisticated behaviors. 

\subsection{Sim to Real Transfer}

\emph{Domain adaption} methods~\citep{DBLP:journals/corr/TzengDHFPLSD15, DBLP:journals/corr/GuptaDLAL17}, progressive nets~\citep{DBLP:conf/corl/RusuVRHPH17}, and learning inverse dynamics models~\citep{DBLP:journals/corr/ChristianoSMSBT16} were all proposed to help with sim to real transfer.
All of these methods assume access to real data.
An alternative approach is to make the policy itself more adaptive during training in simulation using \emph{domain randomization}.
Domain randomization was used to transfer object pose estimators~\citep{tobin2017domain} and vision policies for fly drones~\citep{DBLP:conf/rss/SadeghiL17}.
This idea has also been extended to dynamics randomization~\citep{DBLP:journals/corr/AntonovaCSK17, DBLP:journals/corr/abs-1804-10332, DBLP:conf/rss/YuTLT17} to learn a robust policy that transfers to similar environments but with different dynamics.
Domain randomization was also used to plan robust grasps~\citep{DBLP:conf/rss/MahlerLNLDLOG17, DBLP:journals/corr/abs-1709-06670, DBLP:journals/corr/abs-1710-06425} and to transfer learned locomotion~\citep{DBLP:journals/corr/abs-1804-10332} and grasping~\citep{DBLP:journals/corr/abs-1802-09564} policies for relatively simple robots.
Pinto et al.~\citep{DBLP:conf/icml/PintoDSG17} propose to use \emph{adversarial training} to obtain more robust policies and show that it also helps with transfer to physical robots~\citep{DBLP:conf/icra/PintoDG17}.

\section{Conclusion}
In this work, we demonstrate that
in-hand manipulation skills learned with RL in a simulator can achieve an unprecedented
level of dexterity on  a physical five-fingered hand.
This is possible due to extensive randomizations of the simulator,
large-scale distributed training infrastructure, policies with memory, and
a choice of sensing modalities which can be modelled in the simulator.
Our results demonstrate that,
contrary to a common belief,
contemporary deep RL algorithms
can be applied to solving complex real-world robotics problems which
are beyond the reach of existing non-learning-based approaches.

\section*{Acknowledgements}

We would like to thank Rachel Fong, Ankur Handa and a former OpenAI employee for exploratory work and helpful discussions, a former OpenAI employee for advice and some repairs on hardware and contributions to the low-level PID controller, Pieter Abbeel for helpful discussions, Gavin Cassidy and Luke Moss for their support in maintaining the Shadow hand, and everybody at OpenAI for their help and support.

We would also like to thank the following people for providing feedback on earlier versions of this manuscript: Pieter Abbeel, Joshua Achiam, Tamim Asfour, Aleksandar Botev, Greg Brockman, Rewon Child, Jack Clark, Marek Cygan, Harri Edwards, Ron Fearing, Ken Goldberg, Anna Goldie, Edward Mehr, Azalia Mirhoseini, Lerrel Pinto, Aditya Ramesh, Ian Rust, John Schulman, Shubho Sengupta, and Ilya Sutskever.

\medskip
{
\small
\bibliography{paper}
}

\newpage
\appendix
\appendixpage
\startcontents[appendices]
\printcontents[appendices]{l}{1}{\setcounter{tocdepth}{2}}
\newpage

\section{Reinforcement Learning Background}\label{sec:rl}

\subsection{Reinforcement Learning (RL)}

We consider the standard reinforcement learning formalism
consisting of an agent interacting with an environment.
To simplify the exposition we assume in this section that the environment is fully observable.\footnote{The environments
we consider in the paper are only partially observable.}
An environment
is described by
a set of states $\S$,
a set of actions $\A$,
a distribution of initial states $p(s_0)$,
a reward function $r : \S \times \A \rightarrow \R$,
transition probabilities $p(s_{t+1}|s_t,a_t)$,
and a discount factor $\gamma \in [0,1]$.

A policy $\pi$ is a mapping from state to a distribution over actions.
Every episode starts by sampling an initial state $s_0$.
At every timestep $t$ the agent produces an action based on the current state:
$a_t \sim \pi(\cdot|s_t)$.
In turn, the agents receives a reward $r_t=r(s_t,a_t)$ and the environment's new state $s_{t+1}$, which is sampled from the distribution $p(\cdot|s_t,a_t)$.
The discounted sum of future rewards, also referred to as the \emph{return}, is defined as
$R_t=\sum_{i=t}^\infty \gamma^{i-t} r_i$.
The agent's goal is to maximize its expected return $\E [R_0|s_0]$, where
the expectation is taken over the initial state distribution, policy, and environment transitions accordingly to the dynamics
specified above.
The \emph{Q-function} or \emph{action-value} function is defined as $Q^\pi(s_t,a_t)=\E[R_t|s_t,a_t]$, while the
\emph{V-function} or \emph{state-value} function is defined as $V^\pi(s_t)=\E[R_t|s_t]$.
The value $A^\pi(s_t,a_t)=Q^\pi(s_t,a_t)-V^\pi(s_t)$ is called
the \emph{advantage} and tells whether the action $a_t$ is better or worse than an average
action the policy $\pi$ takes in the state~$s_t$.

\subsection{Generalized Advantage Estimator (GAE)} \label{sec:gae}

Let $V$ be an approximator to the value function of some policy, i.e. $V \approx V^\pi$.
The value $$\hat{V}_t^{(k)}=\sum_{i=t}^{t+k-1} \gamma^{i-t} r_i + \gamma^{k} V(s_{t+k}) \approx V^\pi(s_t,a_t)$$
is called the $k$-step return estimator.
The parameter $k$ controls the bias-variance tradeoff of the estimator
with bigger values resulting in an estimator closer
to empirical returns and having less bias and more variance.
\emph{Generalized Advantage Estimator (GAE)} \citep{gae}
is a method of combining multi-step returns in the following way:
$$\hat{V}_t^\text{GAE} = (1-\lambda) \sum_{k>0}\lambda^{k-1} \hat{V}_t^{(k)} \approx V^\pi(s_t,a_t),$$
where $0<\lambda<1$ is a hyperparameter. Using these to estimate the \emph{advantage}:
$$\hat{A}_t^\text{GAE} = \hat{V}_t^\text{GAE} - V(s_t) \approx A^\pi(s_t,a_t).$$
It is possible
to compute the values of this estimator for all states
encountered in an episode in linear time \citep{gae}.

\subsection{Proximal Policy Optimization (PPO)} \label{sec:ppo}

\emph{Proximal Policy Optimization (PPO)} \citep{ppo} is one of the most popular on-policy RL algorithms.
It simultaneously optimizes a stochastic policy as well as an approximator to the value function.
PPO interleaves the collection of new episodes with policy optimization.
After a batch of new transitions is collected, optimization is performed
with minibatch stochastic gradient descent to maximize the objective $$L_{\text{PPO}}=\E \min \left( \frac{\pi(a_t|s_t)}{\pi_{\text{old}}(a_t|s_t)} \hat{A}^\text{GAE}_t,\, \mbox{clip}\left(\frac{\pi(a_t|s_t)}{\pi_{\text{old}}(a_t|s_t)},\,1-\epsilon,\,1+\epsilon \right)\hat{A}^\text{GAE} _t\right),$$ where
$\frac{\pi(a_t|s_t)}{\pi_{\text{old}}(a_t|s_t)}$ is the ratio of the probability of taking the given action under the current policy $\pi$ to the probability
of taking the same action under the old behavioral policy that was used to generate the data,
and $\epsilon$ is a hyperparameter (usually $\epsilon \approx 0.2$).
This loss encourages the policy to take actions which are better than average (have positive advantage)
while clipping discourages bigger changes to the policy by limiting how much can be gained
by changing the policy on a particular data point.
The value function approximator is trained with supervised learning with the target for $V(s_t)$ being $\hat{V}_t^\text{GAE}$.
To boost exploration, it is a common practice to encourage the policy to have high entropy of actions by including an entropy bonus in the optimization objective.
\section{Hardware Description}
\label{app:hardware}

\subsection{ShadowRobot Dexterous Hand}

We use the ShadowRobot Dexterous Hand. Concretely, we use the version with electric motor actuators, EDC hand (EtherCAT-Dual-CAN).

The hand has \num{24} degrees of freedom between its links and is actuated by \num{40} Spectra tendons controlled by \num{20} DC motors in the base of the hand, each actuating a pair of agonist--antagonist tendons connected via a spool.
\num{16} degrees of freedom can be controlled independently whereas the remaining \num{8} joints (which are the joints between the non-thumb finger proximal, middle, and distal segments) form $4$ pairs of coupled joints.

\subsection{PhaseSpace Visual Tracking}
We use a 3D tracking system to localize the tips of the fingers, to perform calibration procedures, and as ground truth for the RGB image-based tracking. The PhaseSpace Impulse X2E tracking system uses active LED markers that blink to transmit a unique ID code and linear detector arrays in the cameras to detect the positions and IDs. The system features capture speeds of up to 960 Hz and positional accuracies of below 20 $\mu$m. The data is exposed as a 3D point cloud together with labels associating the points with stable numerical IDs. Our setup uses 16 cameras distributed spherically around the hand and centered on the palm with a radius of approximiately $0.8$ meters.

\subsection{RGB Cameras} \label{app:cameras}
\begin{figure}[t]
    \centering
    \includegraphics[width=0.9\textwidth]{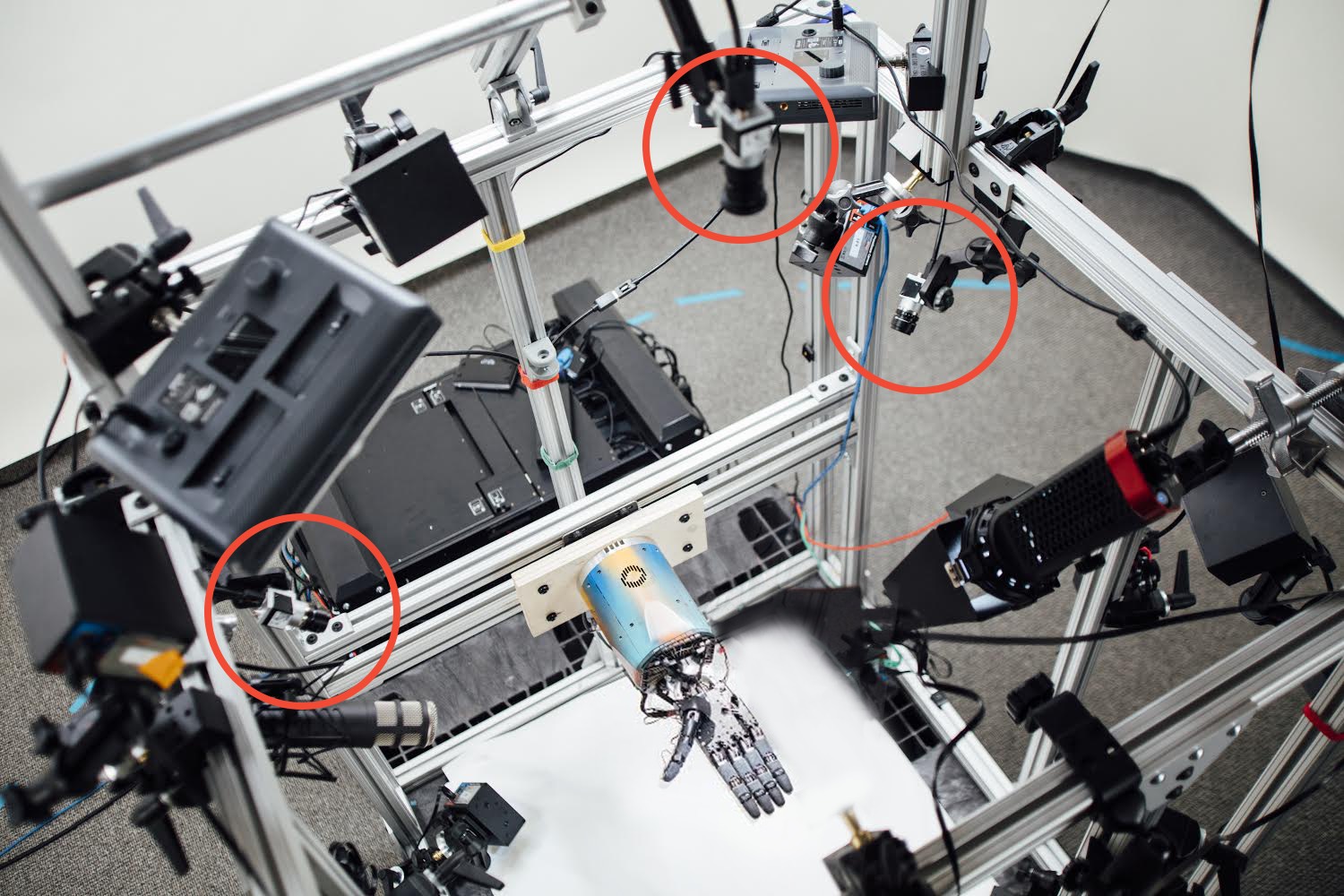}
    \caption{Our 3-camera setup for vision-based state estimation.}
    \label{fig:camera-setup}
\end{figure}

We also use RGB images to track the objects for manipulation. We perform object pose estimation using 3 Basler acA640-750uc RGB cameras with a resolution of 640x480 placed approximately 50 cm from the Shadow hand. We use 3 cameras to resolve pose ambiguities that may occur with monocular vision. We chose these cameras for their flexible parameterization and low latency. Figure \ref{fig:camera-setup} shows the placement of the cameras relative to the hand.

\subsection{Control}

The high-level controller is implemented as a Python program
running a neural network policy implemented in Tensorflow~\citep{tensorflow}
on a GPU. Every 80ms it queries the Phasespace sensors and then runs inference with the neural network to obtain the action, which takes roughly 25ms. The policy outputs an action that specifies the change of position for each actuator, relative to the current position of the joints controlled by the actuator. It then sends the action to the low-level controller.

The low-level controller is implemented in C++ as a separate process on a different machine which is 
connected to the Shadow hand via an Ethernet cable. The controller is written as a real-time system 
--- it is pinned to a CPU core, has preallocated memory, and does not depend on any garbage collector to avoid non-deterministic delays.
The controller receives the relative action, converts it to 
an absolute joint angle and clips to the valid range, then sets each component of the action as the target for a PD controller. Every 5ms, the PD controller queries the Shadow hand joint angle sensors, then attempts to achieve the desired position.

Surprisingly, decreasing time between actions to 40ms increased training time but did not noticeably improve performance in the real world.

\subsection{Joint Sensor Calibration} 
\label{app:sensor-calibration}
The hand contains 26 Hall effect sensors that sense magnetic field rotations along the joint axis.
To transform the raw magnetic measurements from the Hall sensors into joint angles, we use a piecewise linear function interpolated from 3-5 truth points per joint. To calibrate this function, we initialize to the factory default created using physical calibration jigs. For further accuracy, we attach PhaseSpace markers to the fingertips, and minimize error between the position reported by the PhaseSpace markers and the position estimated from the joint angles.
We estimate these linear functions by minimizing reprojection error with \texttt{scipy.minimize}.
\section{Simulated Environment}
\subsection{Deterministic Environment} \label{app:sim}

\paragraph{Simulation.} Our environment is based on the OpenAI Gym robotics environments described in~\citep{plappert2018multi}. We use MuJoCo for simulation~\citep{MuJoCo}.

\paragraph{States.}
The state of the system is \num{60}-dimensional and consists of angles and velocities of all robot joints as well as the position, rotation and velocities (linear and angular) of the object.
Initial states are sampled by placing the object on the palm in a random orientation
and applying random actions for $100$ steps (we discard the trial if the object is dropped in the meantime).

\paragraph{Goals.}
The goal is the desired orientation of the object represented as a quaternion.
A new goal is generated after the current one has been achieved within a tolerance of \SI{0.4}{\radian}.\footnote{I.e. we consider a goal as achieved if there exists a
rotation of the object around an arbitrary axis with an angle smaller than \SI{0.4}{\radian}
which transforms the current orientation into the desired one.}

\paragraph{Observations.}
Described in \autoref{table:policy-inputs}.

\paragraph{Rewards.} \label{sec:reward}
The reward given at timestep $t$ is $r_t=d_t-d_{t+1}$, where
$d_t$ and $d_{t+1}$ are the rotation angles between the desired 
and current object orientations before and after the transition, respectively.
We give an additional reward of $5$ whenever a goal is achieved with the tolerance of \SI{0.4}{\radian}
(i.e. $d_{t+1}<0.4$) and a reward of $-20$ (penalty) whenever the object is dropped.

\paragraph{Actions.}
Actions are \num{20}-dimensional and correspond to the desired angles of the hand joints.
We discretize each action coordinate into $11$ bins of equal size.
Due to the inaccuracy of joint angle sensors
on the physical hand (see Appendix~\ref{app:hardware}), actions are specified
relative to the current hand state.
In particular, the torque applied to the given joint in simulation is equal to $P*(s_t+a-s_{t'})$, where
$s_{t}$ is the joint angle at the time when the action was specified,
$a$ is the corresponding action coordinate, 
$s_{t'}$ is the current joint angle,
 and $P$ is the proportionality coefficient.
For the coupled joints, the desired and actual positions represent the sum of the two joint angles.
All actions are rescaled to the range $[-1,1]$.
To avoid abrupt changes to the action signal, which could harm a physical robot,
we smooth the actions using an exponential moving average\footnote{We use a coefficient of $0.3$ per 80ms.}
before applying them (both in simulation and during deployments on the physical robot).

\paragraph{Timing.}
Each environment step corresponds to \SI{80}{\ms} of real time and consists of \num{10} consecutive MuJoCo steps, each corresponding to \SI{8}{\ms}.
The episode ends when either the policy achieves \num{50} consecutive goals, the policy fails
to achieve the current goal within \num{8} seconds of simulated time, or the object is dropped.

\subsection{Randomizations}
\label{app:randomizations}

A variety of randomizations are applied to the simulator, shrinking the reality gap between the simulated environment and the physical world in order to learn a policy that generalizes to reality.

\paragraph{Physical parameters.}
The physical parameters are sampled at the beginning of every episode and held fixed for the whole episode. The full set of randomized values are displayed in~\autoref{table:rand-physics}.

\paragraph{Observation noise.}
We use two types of noise ---
\textit{correlated} noise which is sampled once per episode and kept fixed,
and an \textit{uncorrelated} Gaussian one.
Apart from Gaussian correlated noise, we also add more structured noise
coming from inaccurate placement of the motion capture markers
by computing the observations using slightly misplaced
markers in the simulator.
The configuration of noise levels is described in~\autoref{table:obs-noise}.
The observation noise is only applied to the policy inputs and not to
the value network inputs as the value function is not used during the deployment on the physical system.

\begin{table}
    \footnotesize
    \centering
    \caption{Standard deviation of observation noise.}
    \renewcommand{\arraystretch}{1.3}
    \begin{tabular}{@{}lll@{}}
        \toprule
        \textbf{Measurement} & \textbf{Correlated noise} & \textbf{Uncorrelated noise} \\ \midrule
        fingertips positions & $1$mm & $2$mm \\
        object position & $5$mm & $1$mm \\
        object orientation & $0.1$rad & $0.1$rad \\ \hline
        fingertip marker positions & $3$mm & \\
        hand base marker position & $1$mm & \\
        \bottomrule
    \end{tabular}
\label{table:obs-noise}
\end{table}

\begin{table}
    \footnotesize
    \centering
    \caption{Standard deviation of action noise.}
    \renewcommand{\arraystretch}{1.3}
    \begin{tabular}{@{}ll@{}}
        \toprule
        \textbf{Noise type} & \textbf{Percentage of the action range} \\ \midrule
        uncorrelated additive & 5\% \\
        correlated additive & 1.5\% \\
        uncorrelated multiplicative & 1.5\% \\
        \bottomrule
    \end{tabular}
\label{table:action-noise}
\end{table}

\paragraph{PhaseSpace tracking errors.}
Noise aside, readings of the motion capture markers from the PhaseSpace system might be occasionally unavailable for a short period of time due to instability of the service. To model such error in the simulator, we mask the fingertip markers with a small probability (0.2 per second) for 1 second so that the policy has a chance to learn how to interact with the environment while the system temporarily loses track of some markers. 

Furthermore, the markers might be occluded while in motion, causing a brief delay of readings of some fingertip positions. In the simulator, a small weightless cuboid site\footnote{A site represents a location of interest relative to the body frame in MuJoCo. Also see \url{http://mujoco.org/book/modeling.html\#site}.} is attached to the back of each nail and we consider a marker occluded whenever a collision with the site is detected as another finger or object is getting too close. If a fingertip marker is deemed occluded, we use its last available position readings instead of the current one.

\paragraph{Action noise and delay.}
We add correlated and uncorrelated Gaussian noise to all actions to account for an imperfect actuation 
system.
The detailed noise levels can be found in~\autoref{table:action-noise}.
Moreover, the real system contains many potential sources of delays between the time that observations are sensed and actions are executed,
from network delay to the computation time of the neural network.
Therefore, we introduce a simple model of action delay to the simulator.
At the beginning of every episode we sample for every actuator whether
it is going to be delayed (with probability $0.5$) or not.
The actions corresponding to delayed actuator are delayed by one environment step, i.e. approximately $80$ms.

\paragraph{Timing randomization.}
We also randomize the timing of environment steps.
Every environment step is simulated as $10$ MuJoCo physics simulator steps
with $\Delta t=8\mbox{ms}+\mbox{Exp}(\lambda)$, where 
$\mbox{Exp}(\lambda)$ denotes the exponential distribution
and the coefficient $\lambda$ is once per episode sampled uniformly from the
range $[1250,10000]$.

\paragraph{Backlash model.}
The physical Shadow Dexterous Hand is tendon-actuated which causes a substantial
amount of backlash, while the MuJoCo model assumes direct actuation on the joints.
In order to account for it, we introduce a simple model of backlash which modifies
actions before they are sent to MuJoCo.
In particular, for every joint we have two parameters which
specify the amount of backlash in each direction, and are denoted $\delta_{-1}$ and $\delta_{+1}$,
as well as a time varying variable $s$ denoting the current state of slack.
We obtained the values of $\delta_{-1}, \delta_{+1}$ through
calibration.
At the beginning of every episode we sample the values of $\delta_{-1}, \delta_{+1}$
from the Gaussian distribution centered around the calibrated values with the standard deviation of $0.1$.
Let $a_{\text{in}} \in [-1,\,1]$ be an action specified by the policy.
Our backlash model works as follows: we compute the new value of the slack variable
$s'=[s+a_{\text{in}} \delta_{\textbf{sgn}(a_\text{in})} \Delta t]_{-1}^{+1}$,
compute the scaling factor
$\alpha=1-\left[\frac{|\textbf{sgn}(a_\text{in})-s|}{|s'-s|+\epsilon}\right]_0^1$,
where $\epsilon=10^{-12}$ is a constant used for numerical stability,
and finally multiply the action by $\alpha$:
$a_{\text{out}} = \alpha a_{\text{in}}$.
    
\paragraph{Random forces on the object.}
To represent unmodeled dynamics, we sometimes apply random forces on the object.
The probability $p$ that a random force is applied is sampled at the beginning of the episode from the loguniform distribution between $0.1\%$ and $10\%$.
Then, at every timestep, with probability $p$ we apply a random force
from the $3$-dimensional Gaussian distribution with the standard deviation equal to $1~m/s^2$ times the mass of the object on each coordinate and decay the force with the coefficient of $0.99$ per $80$ms.

\paragraph{Randomized vision appearance.} 
We randomize the visual appearance of the robot and object, as well as lighting and camera characteristics.
The materials and textures are randomized for every visible object in the scene.
We randomize the hue, saturation, and value for the object faces around calibrated values from real-world measurements.
The color of the robot is uniformly randomized. Material properties such as glossiness and shininess are randomized as well. Camera position and orientation are slightly randomized around values calibrated to real-world locations.
Lights are randomized individually, and intensities are scaled based on a randomly drawn total intensity. After rendering the scene to images from the three separate cameras, additional augmentation is applied.
The images are linearly normalized to have zero mean and unit variance.
Then the image contrast is randomized, and finally per-pixel Gaussian noise is added. Details are in~\autoref{table:vision-randomization}.

\begin{savenotes}
\begin{table}
    \footnotesize
    \centering
    \caption{Vision randomizations.}
    \renewcommand{\arraystretch}{1.3}
    \begin{tabular}{@{}ll@{}}
        \toprule
        \textbf{Randomization type} & \textbf{Range} \\ \midrule
        number of cameras & 3 \\
        camera position & $\pm$ 1.5 mm \\
        camera rotation & 0--3$^{\circ}$ around a random axis \\
        camera field of view & $\pm$ 1$^{\circ}$ \\
        \hline
        robot material colors & RGB \\
        robot material metallic level & 5\%--25\%\footnote{In units used by Unity. See \url{https://unity3d.com/learn/tutorials/s/graphics}.} \\
        robot material glossiness level & 0\%--100\%\footnotemark[\value{footnote}] \\
        \hline
        object material hue & calibrated hue $\pm$ 1\% \\
        object material saturation & calibrated saturation $\pm$ 15\% \\
        object material value & calibrated value $\pm$ 15\% \\
        object metallic level & 5\%--15\%\footnotemark[\value{footnote}] \\
        object glossiness level & 5\%--15\%\footnotemark[\value{footnote}] \\
        \hline
        number of lights & 4--6 \\
        light position & uniform over upper half-sphere \\
        light relative intensity & 1--5 \\
        total light intensity & 0--15\footnotemark[\value{footnote}] \\
        \hline
        image contrast adjustment & 50\%--150\% \\
        additive per-pixel Gaussian noise & $\pm$ 10\% \\
        \bottomrule
    \end{tabular}
\label{table:vision-randomization}
\end{table}
\end{savenotes}

\subsection{MuJoCo Model Calibration}
\label{app:model-calibration}

The MuJoCo XML model of the hand requires many parameters, which are then used as the mean of the randomized distribution of each parameter for the environment. Even though substantial randomization is required to achieve good performance on the physical robot, we have found that it is important for the mean of the randomized distributions to correspond to reasonable physical values. We calibrate these values by recording a trajectory on the robot, then optimizing the default value of the parameters to minimize error between the simulated and real trajectory.

To create the trajectory, we run two policies in sequence against each finger. The first policy measures behavior of the joints near their limits by extending the joints of each finger completely inward and then completely outward until they stop moving. The second policy measures the dynamic response of the finger by moving the joints of each finger inward and then outward in a series of oscillations. The recorded trajectory across all fingers lasts a few minutes.

To optimize the model parameters, these trajectories are then replayed as open-loop action sequences in the simulator. 
The optimization objective is to match joint angles after $1$ second of running actions. Parameters 
are adjusted using iterative coordinate descent until the error is minimized. We exclude modifications to the XML
that does not yield improvement over $0.1\%$.

For each joint, we optimize the damping, equilibrium position, static friction loss, stiffness, margin, and the minimum and maximum of the joint range. For each actuator, we optimize the proportional gain, the force range, and the magnitude of backlash in each direction.  Collectively, this corresponds to 264 parameter values.

\FloatBarrier
\section{Optimization Details}\label{app:hyper}
\subsection{Control Policy}\label{app:hyper-ppo}

\begin{figure}[t]
    \begin{minipage}[c]{0.6\textwidth}
        \includegraphics[width=0.9\textwidth]{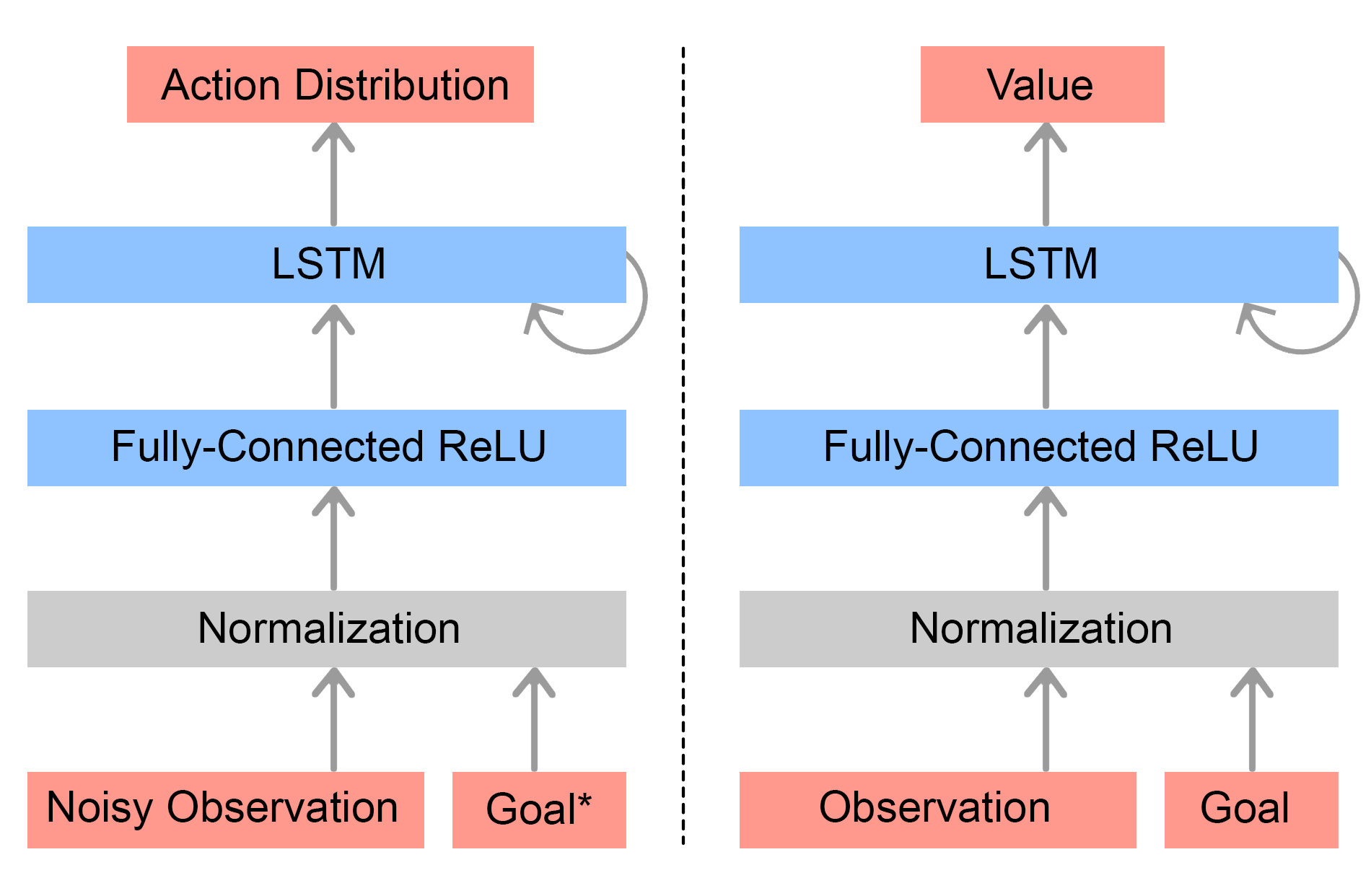}
    \end{minipage}\hfill
    \begin{minipage}[c]{0.4\textwidth}
        \caption{Policy network (left) and value network (right). Each network consists
        of an input normalization, a single fully-connected hidden layer with ReLU activations \citep{relu}
        and a recurrent LSTM block \citep{lstm}. The normalization block
        subtracts the mean value of each coordinate (across all data gathered so far),
        divides by the standard deviation, and removes outliers by clipping.
        There is no weight sharing between the two networks.
        The goal provided to the policy is the noisy relative target orientation (see \autoref{table:policy-inputs} for details).}
        \label{fig:ppo}
    \end{minipage}
\end{figure}

We normalize all observations given to the policy and value networks with running means and standard deviations. We then clip observations such that they are within 5 standard deviations of the mean. We normalize the advantage estimates within each minibatch. We also normalize targets for the value function with running statistics. The network architecture is depicted in \autoref{fig:ppo}.

\begin{table}[h!]
    \footnotesize
    \centering
    \caption{Hyperparameters used for PPO.}
    \renewcommand{\arraystretch}{1.3}
    \begin{tabular}{@{}ll@{}}
        \toprule
        \textbf{Hyperparameter} & \textbf{Value} \\ \midrule
        hardware configuration & 8 NVIDIA V100 GPUs + 6144 CPU cores \\ 
        action distribution & categorical with $11$ bins for each action coordinate \\
        discount factor $\gamma$ & $0.998$ \\
        Generalized Advantage Estimation $\lambda$ & $0.95$ \\
        entropy regularization coefficient & $0.01$ \\
        PPO clipping parameter $\epsilon$ & $0.2$ \\
        optimizer & Adam~\citep{adam} \\
        learning rate & 3e-4 \\
        batch size (per GPU) & $80$k chunks x $10$ transitions = $800$k transitions \\
        minibatch size (per GPU) & $25.6$k transitions \\
        number of minibatches per step & $60$ \\
        network architecture & dense layer with ReLU + LSTM \\
        size of dense hidden layer & 1024 \\
        LSTM size & 512 \\
    \bottomrule\end{tabular}
    \label{tbl:ppo}
\end{table}

\FloatBarrier

\subsection{Vision Model}\label{app:vision-hyper}\label{app:vision_training}
Vision training hyperparameters are given in \autoref{tbl:vision-hyp} and the details of the model architecture are given in \autoref{tbl:vision-hyper-arch}.

We also apply data augmentation for training. More specifically, we leave the object pose as is with $20\%$ probability, rotate the object by $90^{\circ}$ around its main axes with $40\%$ probability, and ``jitter'' the object by adding Gaussian noise to both the position and rotation indepdently with $40\%$ probability.

\begin{table}[h!]
    \centering
    \footnotesize
    \caption{Hyperparameters used for the vision model training.}
    \renewcommand{\arraystretch}{1.3}
    \begin{tabular}{@{}ll@{}}
        \toprule
        \textbf{Hyperparameter} & \textbf{Value} \\ \midrule
        hardware configuration & 3 NVIDIA P40 GPUs\footnote{Two GPUs are used for rendering and one for the optimization.} + 32 CPU cores  \\
        optimizer & Adam~\citep{adam} \\
        learning rate & $0.0005$, halved every $20\,000$ batches \\
        minibatch size &  $64 \times 3 = 192$ RGB images \\
        image size & $200 \times 200$ pixels \\
        weight decay regularization & $0.001$ \\
        number of training batches & $400\,000$ \\
        network architecture & shown in \autoref{fig:vision-architecture} \\
    \bottomrule\end{tabular}
    \label{tbl:vision-hyp}
\end{table}

\begin{table}[h!]
    \centering
    \footnotesize
    \caption{Hyperparameters for the vision model architecture.}
    \begin{tabular}{@{}ll@{}}
        \toprule
        \textbf{Layer} & \textbf{Details} \\ \midrule
        Input RGB Image & $200\times200\times3$ \\
        Conv2D & 32 filters, $5\times5$, stride 1, no padding \\
        Conv2D & 32 filters, $3\times3$, stride 1, no padding \\
        Max Pooling & $3\times3$, stride 3 \\
        ResNet & 1 block, 16 filters,  $3\times3$, stride 3 \\
        ResNet & 2 blocks, 32 filters, $3\times3$, stride 3 \\
        ResNet & 2 blocks, 64 filters, $3\times3$, stride 3 \\
        ResNet & 2 blocks, 64 filters, $3\times3$, stride 3 \\
        Spatial Softmax & \\
        Flatten & \\
        Concatenate & all 3 image towers combined\\ \midrule
        Fully Connected & 128 units \\
        Fully Connected & output dimension ($3$ position + $4$ rotation) \\
    \bottomrule\end{tabular}
    \label{tbl:vision-hyper-arch}
\end{table}

 \FloatBarrier

\end{document}